\newif\ifarxiv
\arxivfalse

\documentclass{article}

\usepackage{microtype}
\usepackage{graphicx}
\usepackage{subfigure}
\usepackage{booktabs} 

\usepackage{wrapfig}

\usepackage{hyperref}

\usepackage[accepted]{icml2024}

\usepackage{amsmath}
\usepackage{amssymb}
\usepackage{mathtools}
\usepackage{amsthm}

\usepackage[capitalize,noabbrev]{cleveref}

\theoremstyle{plain}

\theoremstyle{definition}

\theoremstyle{remark}

\hypersetup{colorlinks,linkcolor={blue},citecolor={magenta},urlcolor={blue}}  

\usepackage[textsize=tiny]{todonotes}

\newcommand{\moe}{$\textrm{Top1-MoE}$}
\newcommand{\softmoe}{$\textrm{Soft MoE}$}
\newcommand{\token}[1]{$\textrm{#1}$}

\newcommand{\mdp}[1]{\mathcal{#1}}

\icmltitlerunning{Mixtures of Experts Unlock Parameter Scaling for Deep RL}

\begin{document}

\twocolumn[
\icmltitle{Mixtures of Experts Unlock Parameter Scaling for Deep RL}
    
    
    
    \icmlsetsymbol{equal}{*}
    
    \begin{icmlauthorlist}
    \icmlauthor{Johan Obando-Ceron*}{dm,mila,udem}
    \icmlauthor{Ghada Sokar*}{dm}
    \icmlauthor{Timon Willi*}{dm,ox}
    \icmlauthor{Clare Lyle}{dm}
    \icmlauthor{Jesse Farebrother}{dm,mila,mcgill}
    \icmlauthor{Jakob Foerster}{ox}
    \icmlauthor{Karolina Dziugaite}{dm}
    \icmlauthor{Doina Precup}{dm,mila,mcgill}
    \icmlauthor{Pablo Samuel Castro}{dm,mila,udem}
    \end{icmlauthorlist}
    
    \icmlaffiliation{mila}{Mila - Québec AI Institute}
    \icmlaffiliation{mcgill}{McGill University}
    \icmlaffiliation{udem}{Université de Montréal}
    \icmlaffiliation{dm}{Google DeepMind}
    \icmlaffiliation{ox}{University of Oxford}

    \icmlcorrespondingauthor{Johan Obando-Ceron}{jobando0730@gmail.com}
    \icmlcorrespondingauthor{Pablo Samuel Castro}{psc@google.com}

    \icmlkeywords{Machine Learning, ICML}

    \vskip 0.3in
]



\printAffiliationsAndNotice{\icmlEqualContribution} 

\ifarxiv
\else
    
\begin{abstract}
The recent rapid progress in (self) supervised learning models is in large part predicted by empirical scaling laws: a model's performance scales proportionally to its size. Analogous scaling laws remain elusive for reinforcement learning domains, however, where increasing the parameter count of a model often hurts its final performance. In this paper, we demonstrate that incorporating Mixture-of-Expert (MoE) modules, and in particular \softmoe{}s \citep{puigcerver2023sparse}, into value-based networks results in more parameter-scalable models, evidenced by substantial performance increases across a variety of training regimes and model sizes. This work thus provides strong empirical evidence towards developing scaling laws for reinforcement learning. \href{https://github.com/google/dopamine/tree/master/dopamine/labs/moes}{\textbf{We make our code publicly available.}}

\end{abstract}

\fi

\section{Introduction}
\label{sec:introduction}
Deep Reinforcement Learning (RL) -- the combination of reinforcement learning algorithms with deep neural networks -- has proven effective at producing agents that perform complex tasks at super-human levels \citep{mnih2015humanlevel, berner2019dota, vinyals2019grandmaster, fawzi2022discovering, Bellemare2020AutonomousNO}. While deep networks are critical to any successful application of RL in complex environments, their design and learning dynamics in RL remain a mystery. Indeed, recent work highlights some of the surprising phenomena that arise when using deep networks in RL, often going against the behaviours observed in supervised learning settings \citep{ostrovski2021tandem,kumar2021implicit, lyle2022understanding,graesser2022state,nikishin22primacy,sokar2023dormant,ceron2023small}.

The supervised learning community convincingly showed that larger networks result in improved performance, in particular for language models \citep{kaplan2020scaling}. In contrast, recent work demonstrates that scaling networks in RL is challenging and requires the use of  sophisticated techniques to stabilize learning, such as supervised auxiliary losses, distillation, and pre-training \citep{farebrother2022proto, taiga2022investigating, schwarzer23bbf}. Furthermore, deep RL networks are {\em under-utilizing} their parameters, which may account for the observed difficulties in obtaining improved performance from scale~\citep{kumar2021implicit, lyle2022understanding, sokar2023dormant}. Parameter count cannot be scaled efficiently if those parameters are not used effectively.

\ifarxiv
  \begin{figure*}[!t]
    \centering
    \includegraphics[width=0.6\linewidth]{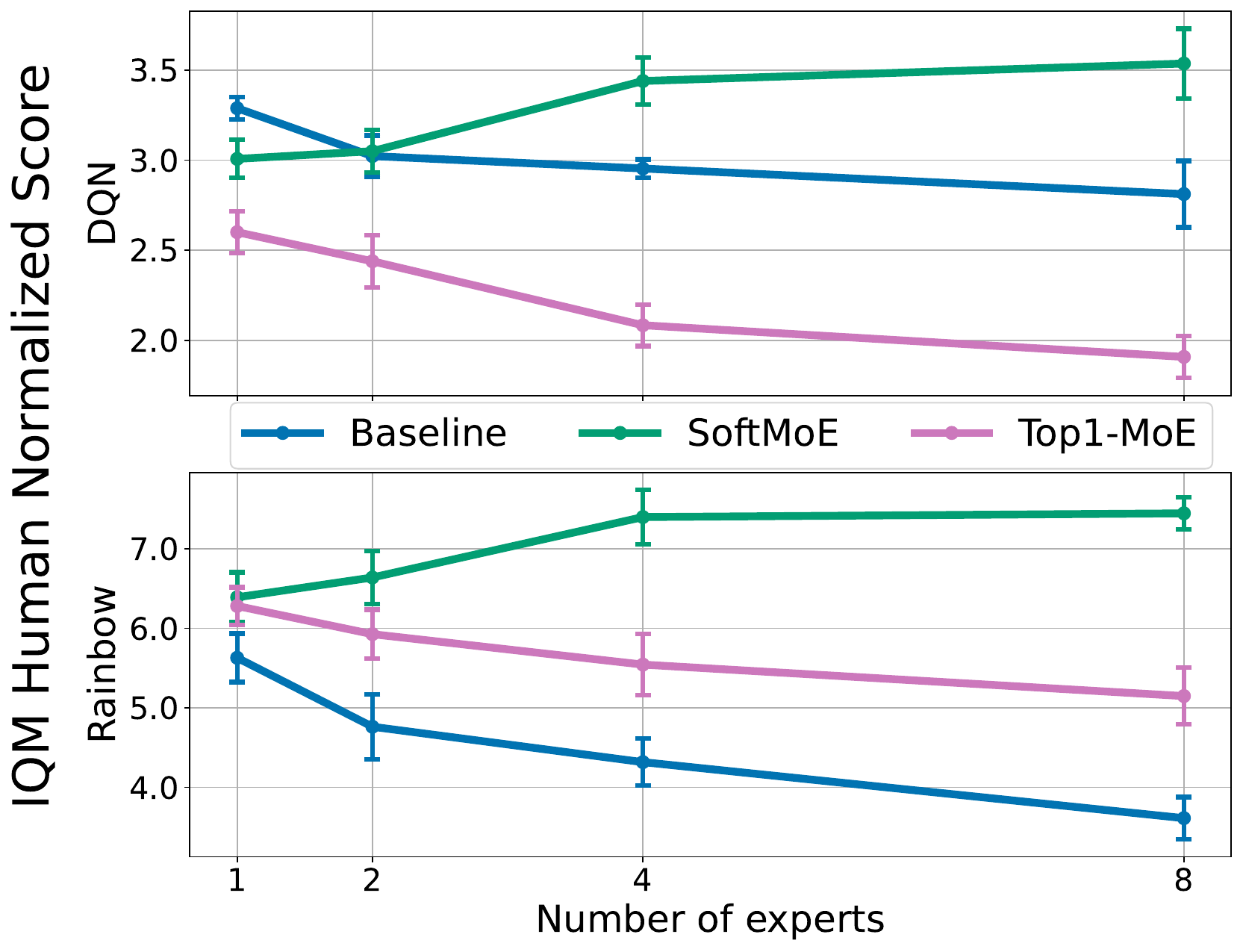}
    \caption{{\bf The use of Mixture of Experts allows the performance of DQN (top) and Rainbow (bottom) to scale with an increased number of parameters}. While \softmoe{} helps in both cases and improves with scale, \moe{} only helps in Rainbow, and does not improve with scale. The corresponding layer in the baseline is scaled by the number of experts to (approximately) match parameters. IQM scores computed over 200M environment steps over $20$ games, with $5$ independent runs each, and error bars showing $95\%$ stratified bootstrap confidence intervals. The replay ratio is fixed to the standard $0.25$.}
    \label{fig:topline}%
  \end{figure*}
\else
  \begin{figure}[!t]
    \centering
    \includegraphics[width=\linewidth]{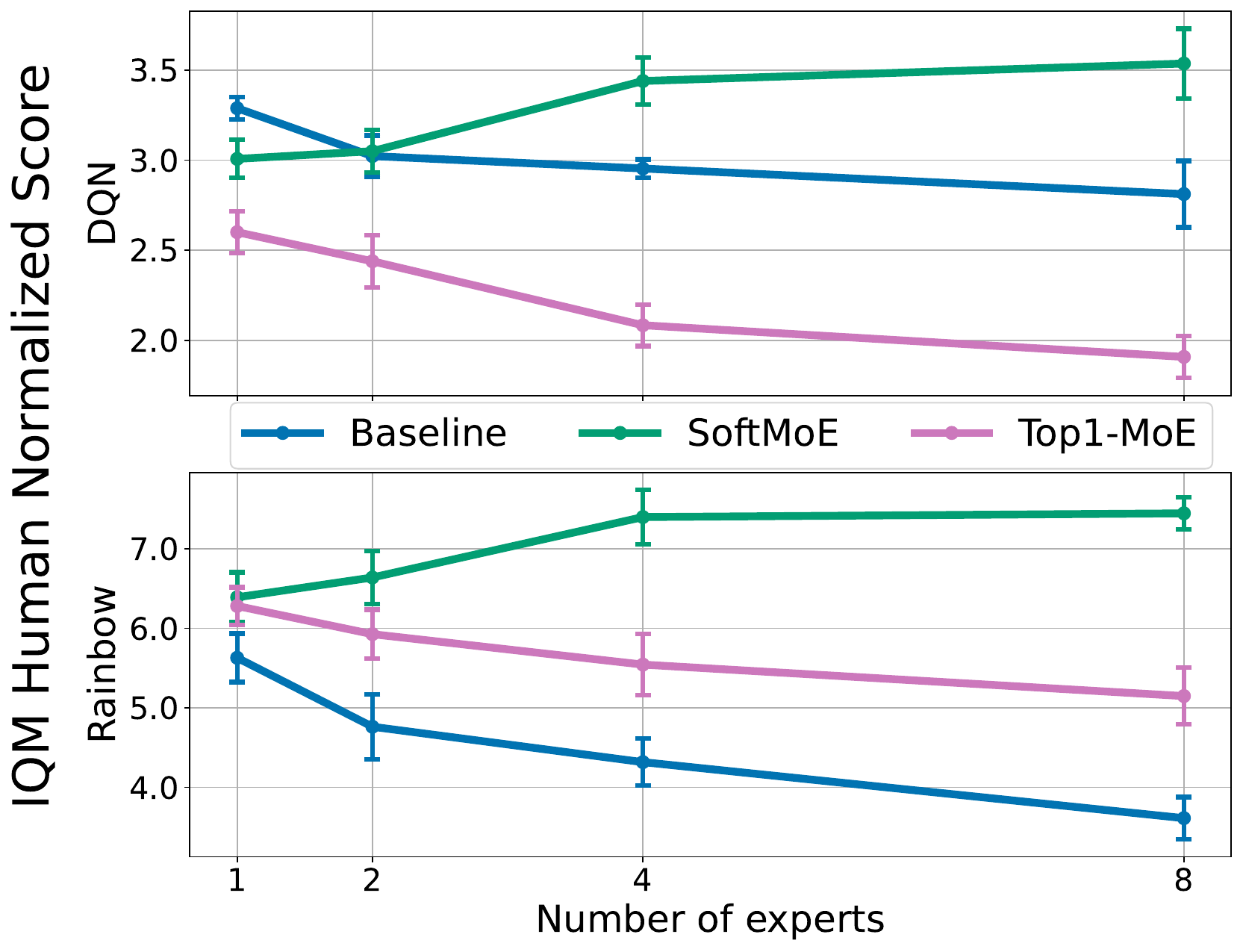}
    \caption{{\bf The use of Mixture of Experts allows the performance of DQN (top) and Rainbow (bottom) to scale with an increased number of parameters}. While \softmoe{} helps in both cases and improves with scale, \moe{} only helps in Rainbow, and does not improve with scale. The corresponding layer in the baseline is scaled by the number of experts to (approximately) match parameters. IQM scores computed over 200M environment steps over 20 games, with 5 independent runs each, and error bars showing 95\% stratified bootstrap confidence intervals. The replay ratio is fixed to the standard $0.25$.}
    \label{fig:topline}%

  \end{figure}
\fi

\ifarxiv
\else
    \begin{figure}[!t]
        \centering
        \includegraphics[width=\linewidth]{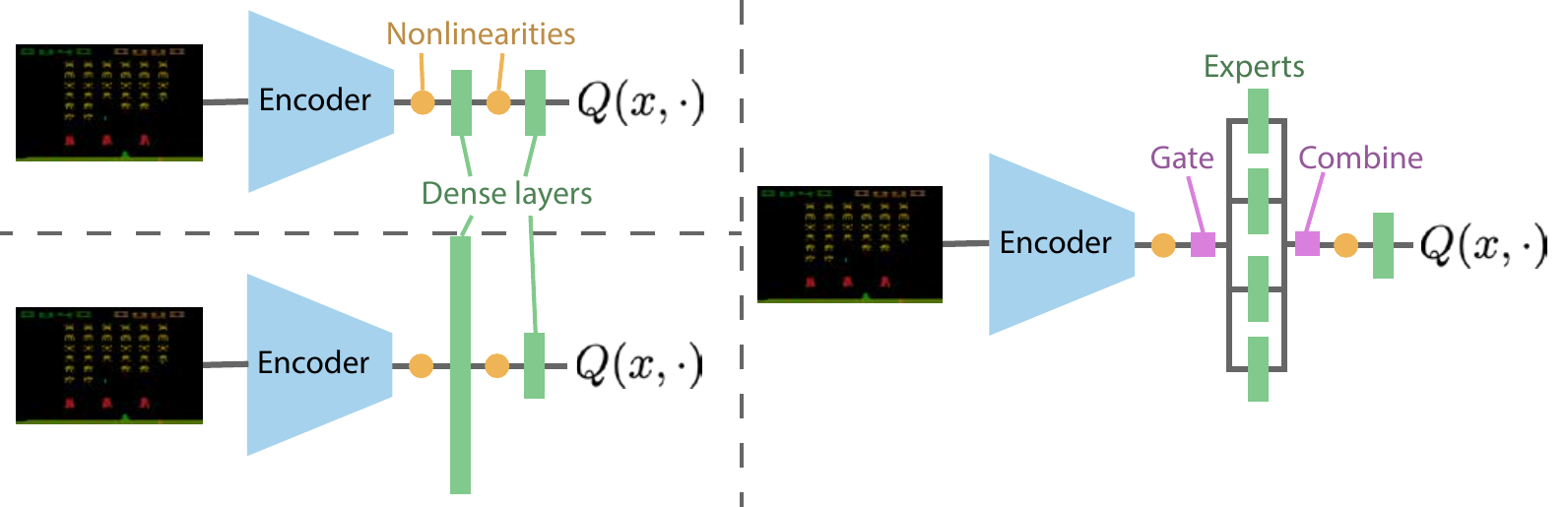}
        \caption{{\bf Incorporating MoE modules into deep RL networks.} {\bf Top left:} Baseline architecture; {\bf bottom left:} Baseline with penultimate layer scaled up; {\bf right:} Penultimate layer replaced with an MoE module.}
        \label{fig:moeArchitecture}%
        \vspace{-0.1cm}
    \end{figure}
\fi

Architectural advances, such as transformers \citep{vaswani2017attention}, adapters \citep{houlsby19parameter}, and Mixtures of Experts \citep[MoEs;][]{shazeer2017outrageously}, have been central to the scaling properties of supervised learning models, especially in natural language and computer vision problem settings. MoEs, in particular, are crucial to scaling networks to billions (and recently trillions) of parameters, because their modularity combines naturally with distributed computation approaches~\citep{fedus2022switch}. Additionally, MoEs induce {\em structured sparsity} in a network, and certain types of sparsity have been shown to improve network performance \citep{evci2020rigl,gale2019state}.

In this paper, we explore the effect of mixture of experts on the parameter scalability of value-based deep RL networks, i.e., does performance increase as we increase the number of parameters? We demonstrate that incorporating \softmoe{}s \citep{puigcerver2023sparse} strongly improves the performance of various deep RL agents, and performance improvements scale with the number of experts used. We complement our positive results with a series of analyses that help us understand the underlying causes for the results in \cref{sec:empiricalResults}. For example, we investigate different gating mechanisms, motivating the use of \softmoe{}, as well as different tokenizations of inputs. Moreover, we analyse different properties of the experts hidden representations, such as dormant neurons~\citep{sokar2023dormant}, which provide empirical evidence as to why \softmoe{} improves performance over the baseline.

Finally, we present a series of promising results that pave the way for further research incorporating MoEs in deep RL networks in \cref{sec:futureDirections}. For instance, in \cref{sec:offline_rl} we show preliminary results that \softmoe{} outperforms the baseline on a set of Offline RL tasks; in \cref{sec:sample_eff}, we evaluate \softmoe{}'s performance in low-data training regimes; and lastly, we show that exploring different architectural designs is a fruitful direction for future research in \cref{sec:arch_exploration}.

\ifarxiv
    \begin{figure}[!h]
        \centering
        \includegraphics[width=\linewidth]{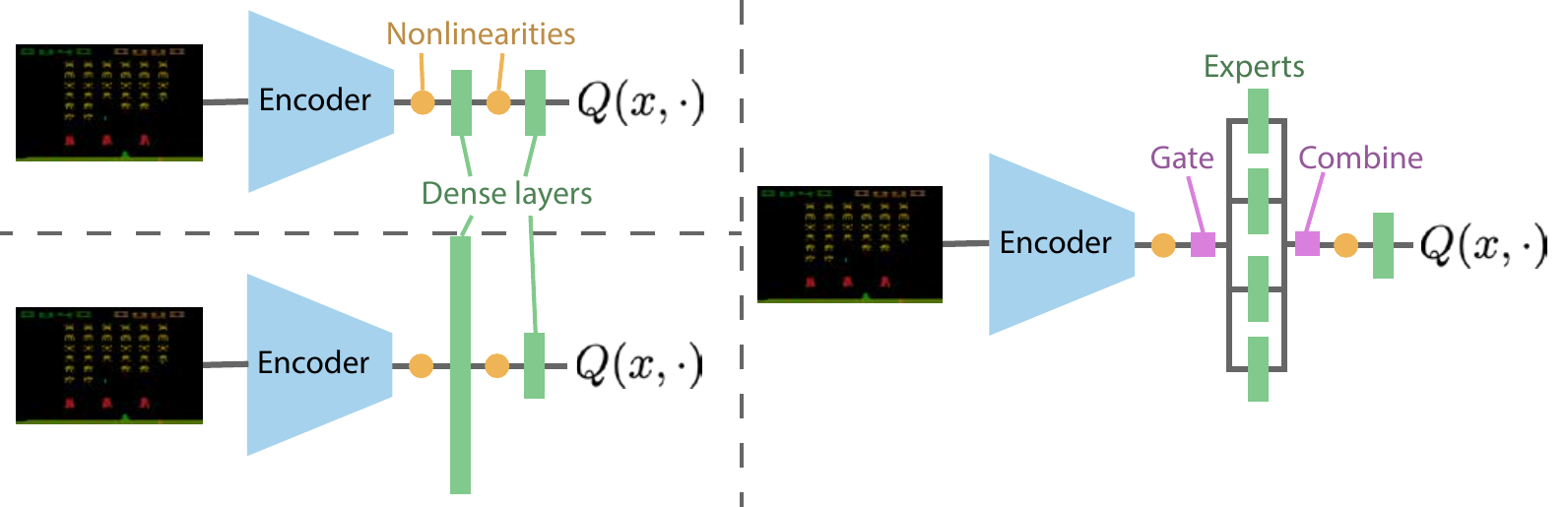}
        \caption{{\bf Incorporating MoE modules into deep RL networks.} {\bf Top left:} Baseline architecture; {\bf bottom left:} Baseline with penultimate layer scaled up; {\bf right:} Penultimate layer replaced with an MoE module.}
        \label{fig:moeArchitecture}%
    \end{figure}
\fi

\begin{figure}[!t]
    \centering
    \includegraphics[width=\linewidth]{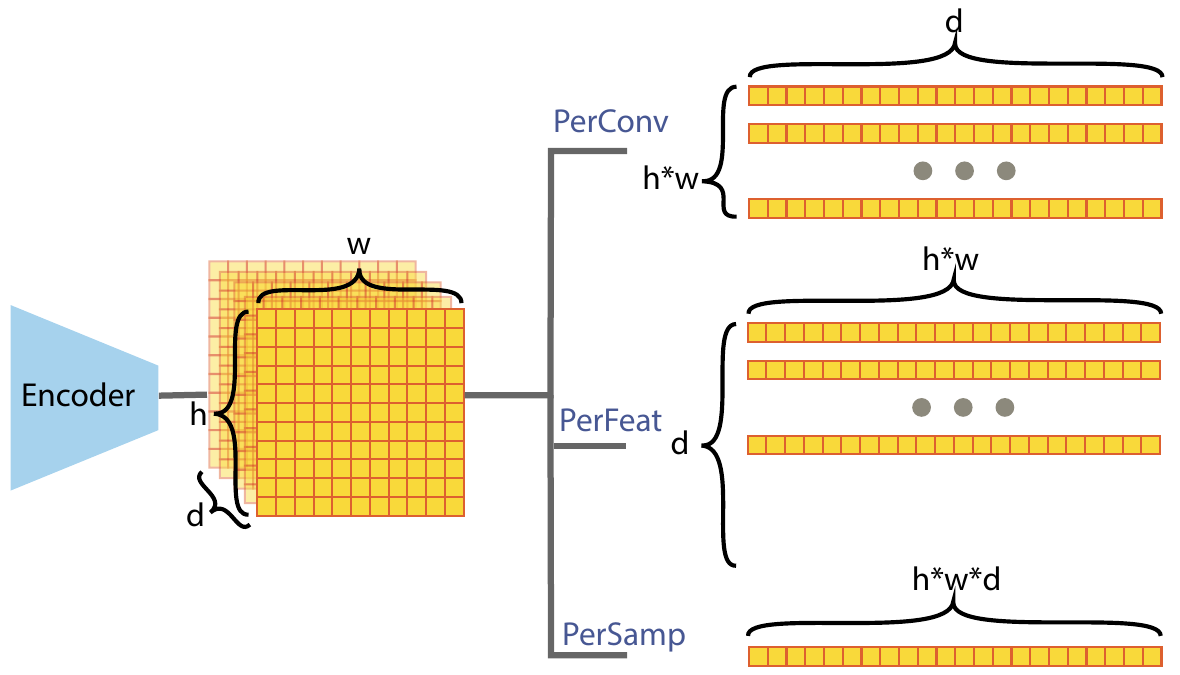}
    \caption{{\bf Tokenization types considered:} PerConv (per convolution), PerFeat (per feature), and PerSamp (per sample).}
    \label{fig:tokenization}%
    \vspace{-0.1cm}
\end{figure}

\section{Preliminaries}
\label{sec:background}

\subsection{Reinforcement Learning}
Reinforcement learning methods are typically employed in sequential decision-making problems, with the goal of finding optimal behavior within a given environment~\citep{sutton98rl}. These environments are typically formalized as Markov Decision Processes (MDPs), defined by the tuple $\langle \mdp{X}, \mdp{A}, \mdp{P}, \mdp{R}, \gamma \rangle$, where $\mdp{X}$ represents the set of states, $\mdp{A}$ the set of available actions, $\mdp{P}:\mdp{X}\times\mdp{A}\to \Delta(\mdp{X})$\footnote{$\Delta(X)$ denotes a distribution over the set $X$.} the transition function, $\mdp{R}:\mdp{X}\times\mdp{A}\to\mathbb{R}$ the reward function, and $\gamma\in [0, 1)$ the discount factor.

An agent's behaviour, or policy, is expressed via a mapping $\pi:\mdp{X}\to\Delta(\mdp{A})$. Given a policy $\pi$, the {\em value} $V^{\pi}$ of a state $x$ is given by the expected sum of discounted rewards when starting from that state and following $\pi$ from then on:\\
$ V^\pi(x) := \underset{\pi, \mdp{P}}{\mathbb{E}}\left[\sum_{t=i}^{\infty} \gamma^t \mdp{R}\left(x_t, a_t \right) \mid x_0 = x \right]. $ The state-action function $Q^{\pi}$ quantifies the value of first taking action $a$ from state $x$, and then following $\pi$ thereafter: $Q^{\pi}(x, a) := \mdp{R}(x, a) + \gamma\underset{x'\sim\mdp{P}(x, a)}{\mathbb{E}}V^{\pi}(x')$. Every MDP admits an optimal policy $\pi^*$ in the sense that $V^{\pi^*} := V^* \geq V^{\pi}$ uniformly over $\mdp{X}$ for all policies $\pi$.

When $\mdp{X}$ is very large (or infinite), function approximators are used to express $Q$, e.g., DQN by \citet{mnih2015humanlevel} uses neural networks with parameters $\theta$, denoted as $Q_{\theta}\approx Q$. The original architecture used in DQN, hereafter referred to as the CNN architecture, comprises of 3 convolutional layers followed by $2$ dense layers, with ReLu nonlinearities \citep{fukushima1969visual} between each pair of layers. In this work, we mostly use the newer Impala architecture \citep{espeholt2018impala}, but will provide a comparison with the original CNN architecture. DQN also used a {\em replay buffer}: a (finite) memory where an agent stores transitions received while interacting with the environment, and samples mini-batches from to compute gradient updates. The {\em replay ratio}\footnote{\label{note1} In the hyperparameters established in \cite{mnih2015humanlevel}, the policy is updated every $4$ environment steps collected, resulting in a replay ratio of $0.25$.} is defined as the ratio of gradient updates to environment interactions, and plays a role in our analyses below. 

Rainbow \citep{Hessel2018RainbowCI} extended the original DQN algorithm with multiple algorithmic components to improve learning stability, sample efficiency, and overall performance. Rainbow was shown to significantly outperform DQN and is an important baseline in deep RL research.

\ifarxiv
\else
    \begin{figure}[!t]
        \centering
        \includegraphics[width=0.245\textwidth]{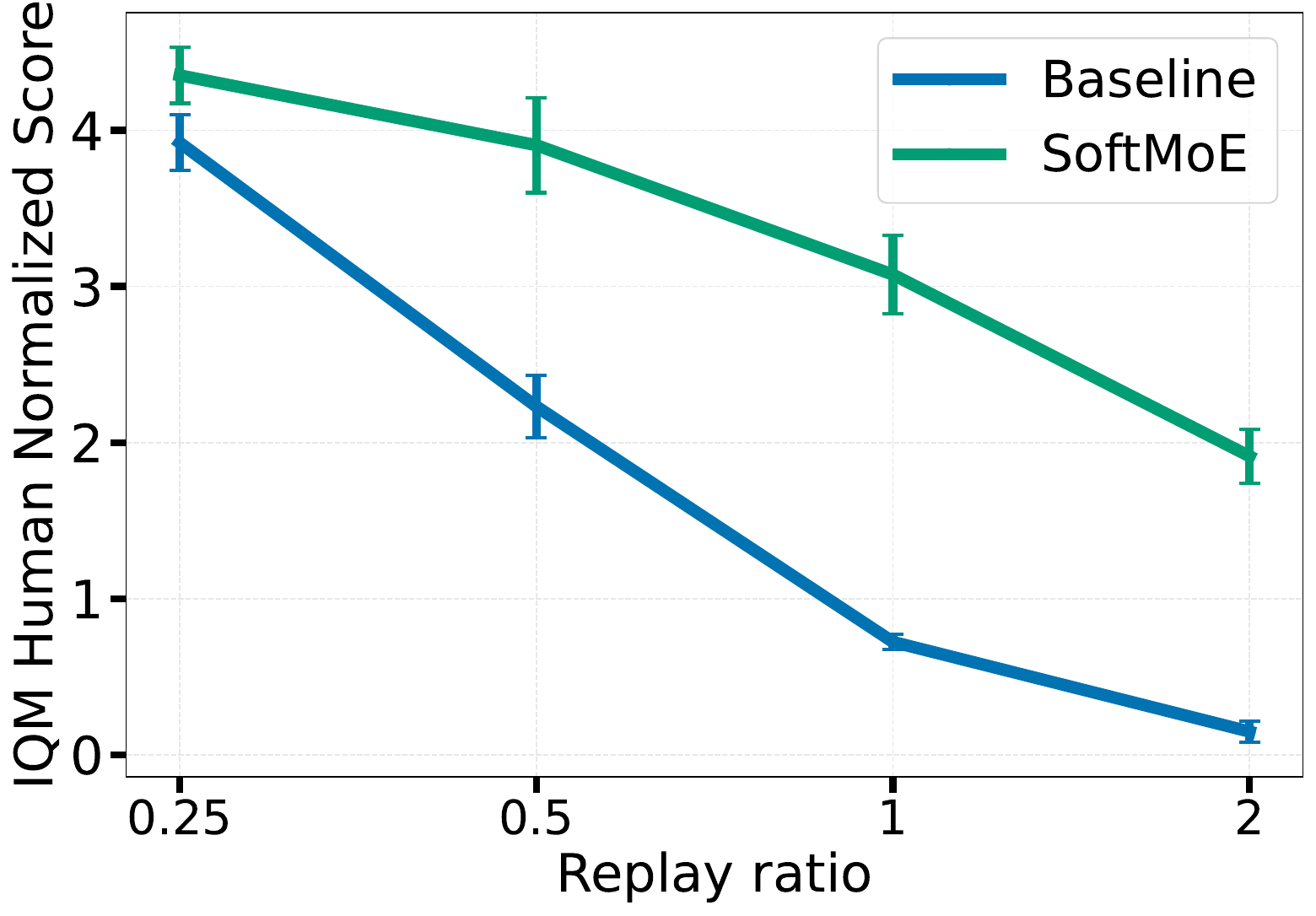}%
        \includegraphics[width=0.235\textwidth]{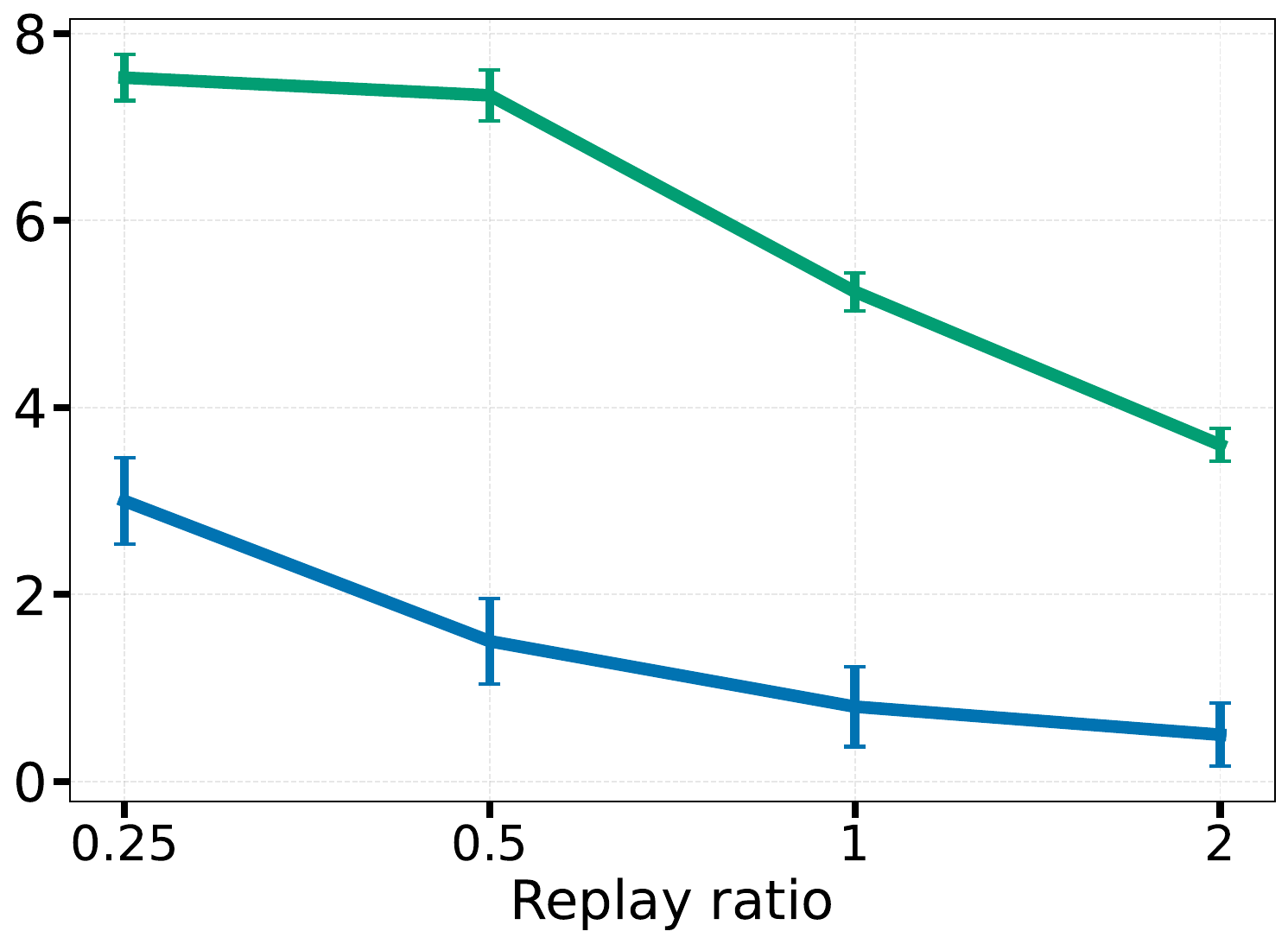}%
        \vspace{-0.4cm}
        \caption{\textbf{\softmoe{} yields performance gains even at high replay ratio values.} DQN (left) and Rainbow (right) with 8 experts. See \cref{sec:setup} for training details.}
        \label{fig:replayRatioPlots}
        \vspace{-0.2cm}
    \end{figure}
\fi

\subsection{Mixtures of Experts}
\label{sec:moeBackground}
Mixtures of experts (MoEs) have become central to the architectures of most modern Large Language Models (LLMs). They consist of a set of $n$ ``expert'' sub-networks activated by a gating network (typically learned and referred to as the {\em router}), which routes each incoming token to $k$ experts \citep{shazeer2017outrageously}. In most cases, $k$ is smaller than the total number of experts (k=1 in our work), thereby inducing sparser activations \citep{fedus2022switch}. These sparse activations enable faster inference and distributed computation, which has been the main appeal for LLM training. MoE modules typically replace the dense feed forward blocks in transformers \citep{vaswani2017attention}. Their strong empirical results have rendered MoEs a very active area of research in the past few years \citep{shazeer2017outrageously,Lewis2021BASELS,fedus2022switch,zhou2022mixture,puigcerver2023sparse,lepikhin2020gshard,zoph2022stmoe,gale2023megablocks}.

Such hard assignments of tokens to experts introduce a number of challenges such as training instabilities, dropping of tokens, and difficulties in scaling the number of experts \citep{fedus2022switch,puigcerver2023sparse}. To address some of these challenges, \citet{puigcerver2023sparse} introduced Soft MoE, which is a fully differentiable soft assignment of tokens-to-experts, replacing router-based hard token assignments. 

Soft assignment is achieved by computing (learned) mixes of per-token weightings for each expert, and averaging their outputs. Following the notation of \citet{puigcerver2023sparse}, let us define the input tokens as $\mathbf{X} \in \mathbb{R}^{m \times d}$, where $m$ is the number of $d$-dimensional tokens. A \softmoe{} layer applies a set of $n$ experts on individual tokens, $\left\{f_i: \mathbb{R}^d \rightarrow \mathbb{R}^d\right\}_{1: n}$. Each expert has $p$ input- and output-slots, represented respectively by a $d$-dimensional vector of parameters.  We denote these parameters by $\boldsymbol{\Phi} \in \mathbb{R}^{d \times(n \cdot p)}$.

\ifarxiv
    \begin{figure}[!t]
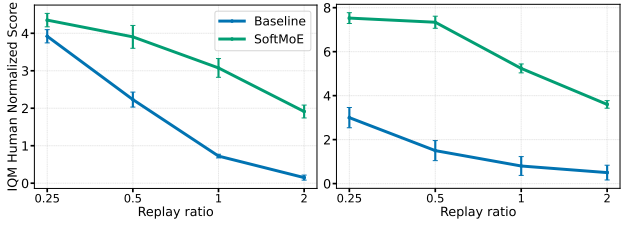

        \centering
        \includegraphics[width=0.245\textwidth]{figures/MOEs_replayratio_dqn_experts8CORR_COLOR2.pdf}%
        \includegraphics[width=0.235\textwidth]{figures/MOEs_replayratio_rainbow_experts8CORR_COLOR2.pdf}%
        \vspace{-0.4cm}
        \caption{\textbf{\softmoe{} yields performance gains even at high replay ratio values.} DQN (left) and Rainbow (right) with 8 experts. See \cref{sec:setup} for training details.}
        \label{fig:replayRatioPlots}
        \vspace{-0.2cm}
    \end{figure}
\fi

The input-slots $\tilde{\mathbf{X}} \in \mathbb{R}^{(n \cdot p) \times d}$ correspond to a weighted average of all tokens: $\tilde{\mathbf{X}}=\mathbf{D}^{\top} \mathbf{X}$, where

\begin{equation}
\begin{gathered}
\mathbf{D}_{i j}=\frac{\exp \left((\mathbf{X} \boldsymbol{\Phi})_{i j}\right)}{\sum_{i^{\prime}=1}^m \exp \left((\mathbf{X} \boldsymbol{\Phi})_{i^{\prime} j}\right)}. \notag
\end{gathered}
\end{equation}

D is typically referred to as the \textit{dispatch weights}. We then denote the expert outputs as $\tilde{\mathbf{Y}}_i=f_{\lfloor i / p\rfloor}\left(\tilde{\mathbf{X}}_i\right)$.
The output of the \softmoe{} layer $\mathbf{Y}$ is the combination of $\tilde{\mathbf{Y}}$ with the \textit{combine weights} $\mathbf{C}$ according to $\mathbf{Y}=\mathbf{C} \tilde{\mathbf{Y}}$, where

\begin{equation}
\begin{gathered}
\mathbf{C}_{i j}=\frac{\exp \left((\mathbf{X} \boldsymbol{\Phi})_{i j}\right)}{\sum_{j^{\prime}=1}^{n \cdot p} \exp \left((\mathbf{X} \boldsymbol{\Phi})_{i j^{\prime}}\right)} \notag.
\end{gathered}
\end{equation}

Note how $\mathbf{D}$ and $\mathbf{C}$ are learned only through $\boldsymbol{\Phi}$, which we will use in our analysis.
The results of \citet{puigcerver2023sparse} suggest that Soft MoE achieves a better trade-off between accuracy and computational cost compared to other MoE methods.
\section{ Mixture of Experts for Deep RL}
\label{sec:moesForDRL}

Below we discuss some important design choices in our incorporation of MoE modules into DQN-based architectures.

\begin{figure}[!t]
    \centering
    \includegraphics[width=0.8\linewidth]{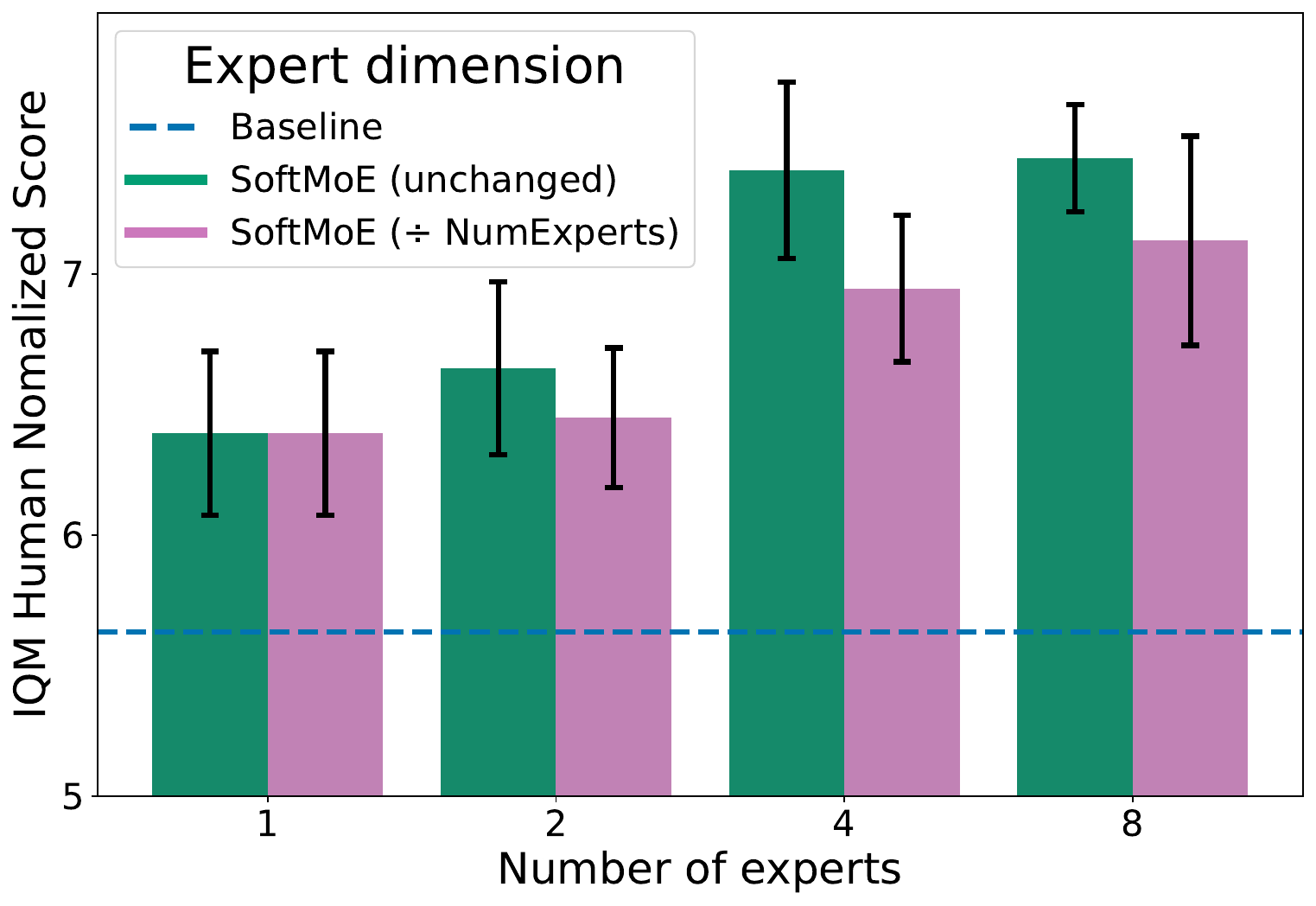}
    \caption{{\bf Scaling down the dimensionality of \softmoe{} experts has no significant impact on performance in Rainbow}. See \cref{sec:setup} for training details.}
    \label{fig:rainbowScaling}%
\end{figure}

\paragraph{Where to place the MoEs?} Following the predominant usage of MoEs to replace dense feed-forward layers \citep{shazeer2017outrageously,fedus2022switch,gale2023megablocks}, we replace the penultimate layer in our networks with an MoE module, where each expert has the same dimensionality as the original dense layer. Thus, we are effectively widening the penultimate layer's dimensionality by a factor equal to the number of experts. \cref{fig:moeArchitecture} illustrates our deep RL MoE architecture. We discuss some alternatives in \cref{sec:arch_exploration}.

\paragraph{What is a token?} MoEs sparsely route inputs to a set of experts \citep{shazeer2017outrageously} and are mostly used in the context of transformer architectures where the inputs are {\em tokens} \citep{fedus2022switch}. For the vast majority of supervised learning tasks where MoEs are used there is a well-defined notion of a token; except for a few works using Transformers \citep{chen2021decisiontransformer}, this is not the case for deep RL networks. Denoting by $C^{(h,w,d)}\in\mathbb{R}^{3}$ the output of the convolutional encoders, we define \emph{tokens} as $d$-dimensional slices of this output; thus, we split $C$ into $h\times w$ tokens of dimensionality $d$ (\token{PerConv} in \cref{fig:tokenization}). This approach to tokenization is taken from what is often used in vision tasks (see the Hybrid architecture of \citet{dosovitskiy2021an}). We did explore other tokenization approaches (discussed in \cref{sec:impactDesign}), but found this one to be the best performing and most intuitive. Finally, a trainable linear projection is applied after each expert to maintain the token size $d$ at the output.

\paragraph{What flavour of MoE to use?} 
We explore the top-$k$ gating architecture of \citet{shazeer2017outrageously} following the simplified $k=1$ strategy of \citet{fedus2022switch}, as well as the \softmoe{} variant proposed by \citet{puigcerver2023sparse}; for the rest of the paper we refer to the former as \moe{} and the latter as \softmoe{}. We focus on these two, as \moe{} is the predominantly used approach, while \softmoe{} is simpler and shows evidence of improving performance. Since \moe{}s activate one expert per token while \softmoe{}s can activate multiple experts per token, \softmoe{}s are arguably more directly comparable in terms of the number of parameters when widening the dense layers of the baseline.

\section{Empirical evaluation}
\label{sec:empiricalResults}

\begin{figure}[!t]
    \centering
    \includegraphics[width=\linewidth]{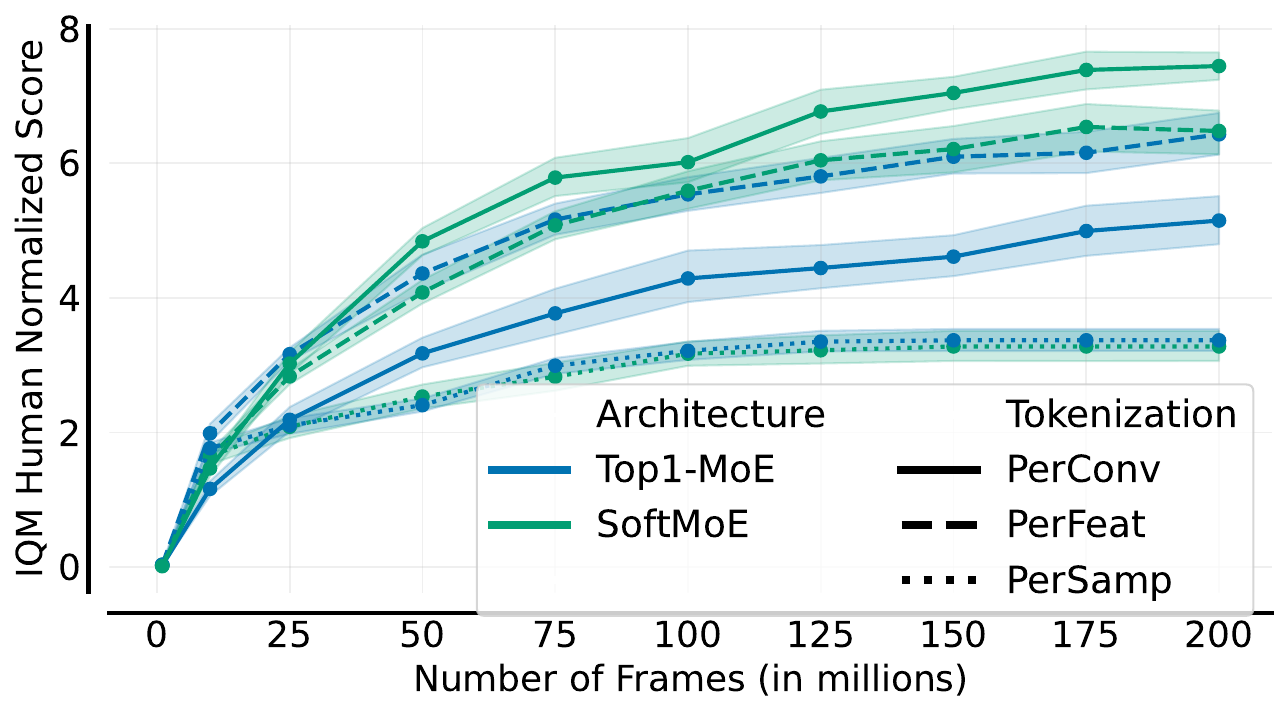}
    \vspace{-0.4cm}
    \caption{{\bf Comparison of different tokenization methods on Rainbow with $8$ experts.} \softmoe{} with PerConv achieves the best results. \moe{} works best with PerFeat. PerSamp is worst over both architectures. See \cref{sec:setup} for training details.}
    \label{fig:rainbowTokenization}%
\end{figure}

\label{sec:setup}
We focus our investigation on DQN \citep{mnih2015humanlevel} and Rainbow \citep{Hessel2018RainbowCI}, two value-based agents that have formed the basis of a large swath of modern deep RL research. Recent work has demonstrated that using the ResNet architecture ~\citep{espeholt2018impala} instead of the original CNN architecture~\citep{mnih2015humanlevel}  yields strong empirical improvements \citep{graesser2022state,schwarzer23bbf}, so we conduct most of our experiments with this architecture. As in the original papers, we evaluate on 20 games from the Arcade Learning Environment (ALE), a collection of a diverse and challenging pixel-based environments \citep{bellemare2012ale}.

Our implementation, included with this submission, is built on the Dopamine library \citep{castro18dopamine}\footnote{Dopamine code available at  \url{https://github.com/google/dopamine}.}, which adds stochasticity (via sticky actions) to the ALE \citep{machado18revisiting}. We use the recommendations from \citet{agarwal2021deep} for statistically-robust performance evaluations, in particular focusing on interquartile mean (IQM). Every experiment was run for 200M environment steps, unless reported otherwise, with $5$ independent seeds, and we report $95\%$ stratified bootstrap confidence intervals. All experiments were run on NVIDIA Tesla P100 GPUs, and each took on average $4$ days to complete.

\begin{figure}[!t]
    \centering
    \includegraphics[width=\linewidth]{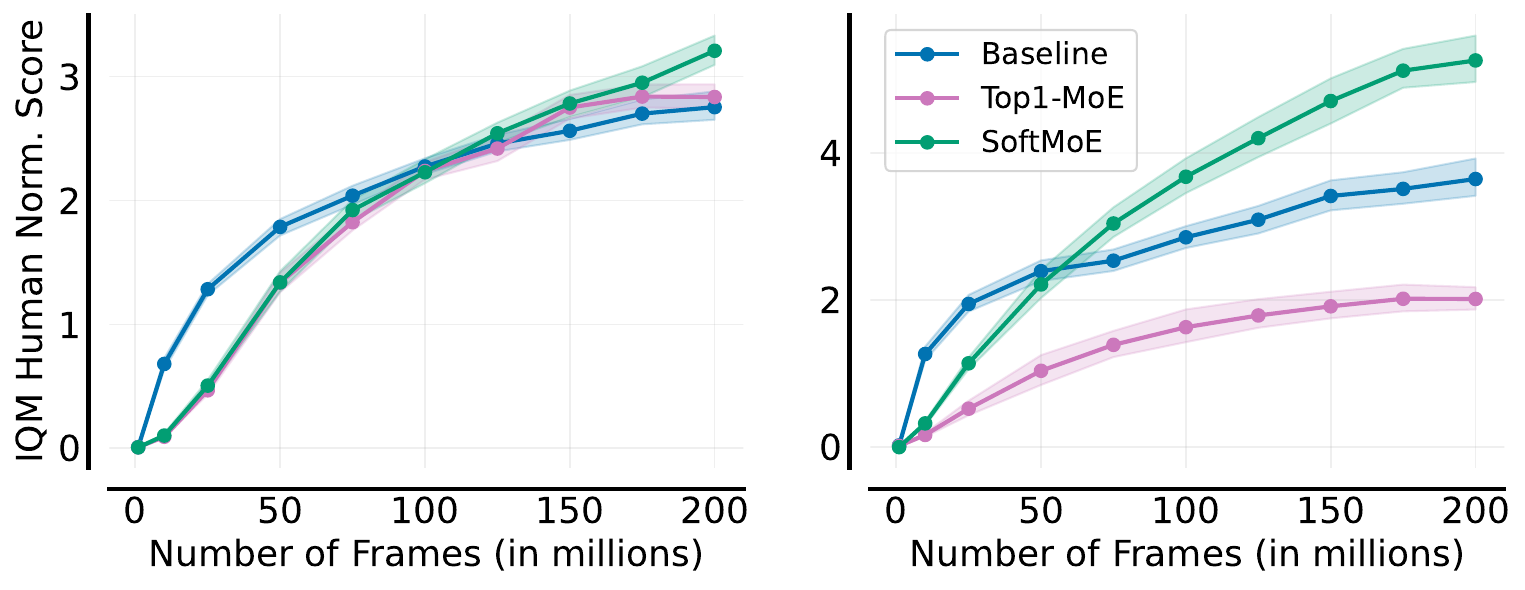}
    \vspace{-0.4cm}
    \caption{{\bf In addition to the Impala encoder, MoEs also show performance gains for the standard CNN encoder on DQN (left) and Rainbow (right)}, with $8$ experts. See \cref{sec:setup} for training details.}
    \label{fig:CNNEval}%
    \vspace{-0.3cm}
\end{figure}

\subsection{\softmoe{} Helps Parameter Scalability}
To investigate the efficacy of \moe{} and \softmoe{} on DQN and Rainbow, we replace the penultimate layer with the respective MoE module (see \cref{fig:moeArchitecture}) and vary the number of experts. Given that each expert is a copy of the original penultimate layer, we are effectively increasing the number of parameters of this layer by a factor equal to the number of experts. To compare more directly in terms of number of parameters, we evaluate simply widening the penultimate layer of the base architectures by a factor equal to the number of experts.

\ifarxiv
\else
    \begin{figure}[!t]
        \centering
        \includegraphics[width=0.24\textwidth]{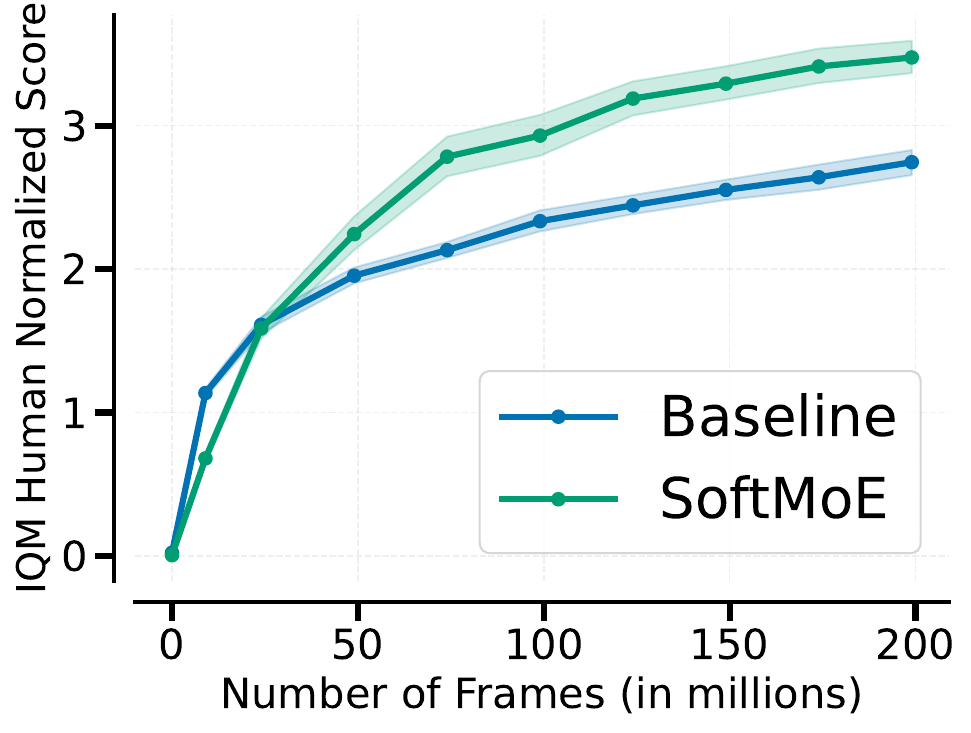}%
        \includegraphics[width=0.24\textwidth]{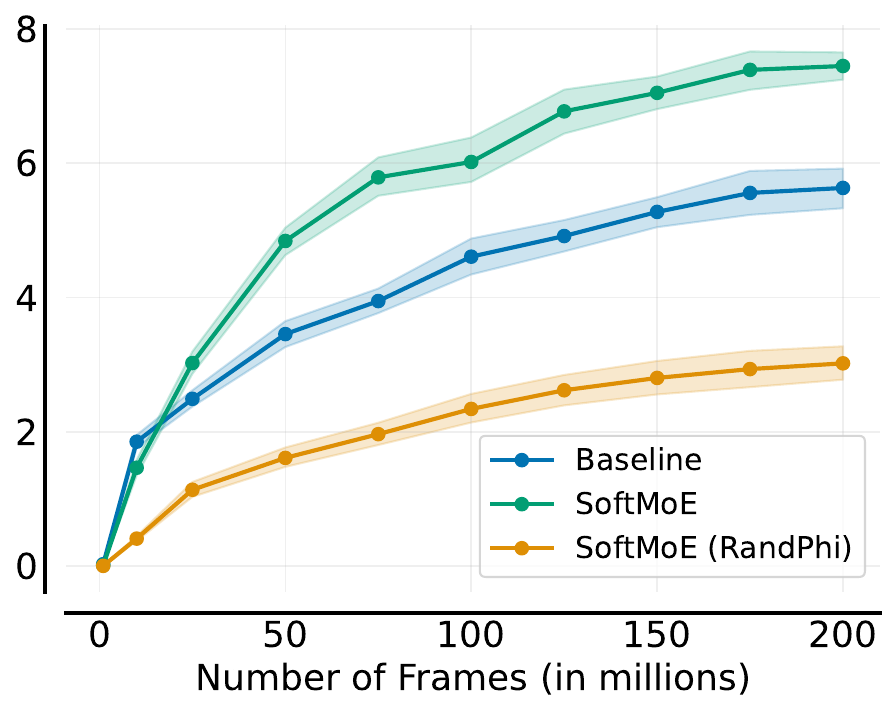}%
        \vspace{-0.4cm}
        \caption{\textbf{(Left) Evaluating the performance on 60 Atari 2600 games; (Right) Evaluating the performance of \softmoe{} with a random $\boldsymbol{\Phi}$ matrix}. (Left) Even over 60 games, \softmoe{} performs better than the baseline. (Right) Learning $\boldsymbol{\Phi}$ is beneficial over a random phi, indicating that \softmoe{}'s perfomance gains are not only due to the distribution of tokens to experts. Both plots run with Rainbow using the Impala architecture and 8 experts. See \cref{sec:setup} for training details.}
        \label{fig:allsuite}
    \end{figure}
\fi

As \cref{fig:topline} demonstrates, \softmoe{} provides clear performance gains, and these gains increase with the number of experts; for instance in Rainbow, increasing the number of experts from $1$ to $8$ results in a $20\%$ performance improvement. In contrast, the performance of the base architectures {\em declines} as we widen its layer; for instance in Rainbow, as we increase the layer multiplier from $1$ to $8$ there is a performance decrease of around $40\%$. This is a finding consistent with prior work demonstrating the difficulty in scaling up deep RL networks \citep{farebrother2022proto, taiga2022investigating, schwarzer23bbf}.
\moe{} seems to provide gains when incorporated into Rainbow, but fails to exhibit the parameter-scalability we observe with \softmoe{}.

It is known that deep RL agents are unable to maintain performance when scaling the {\em replay ratio} without explicit interventions \citep{doro2023sampleefficient,schwarzer23bbf}. In \cref{fig:replayRatioPlots} we observe that the use of \softmoe{}s maintains a strong advantage over the baseline even at high replay ratios, further confirming that they make RL networks more parameter efficient.

\ifarxiv
    \begin{figure}[!t]
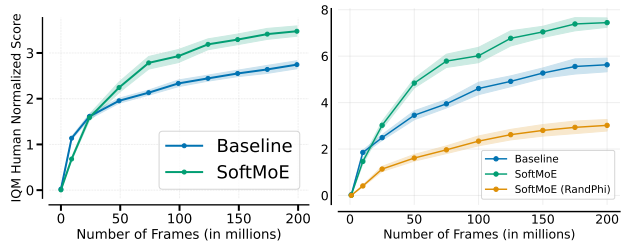

        \centering
        \includegraphics[width=0.24\textwidth]{figures/MOES_all_suite_resumenCORRCOLOR.pdf}%
        \includegraphics[width=0.24\textwidth]{figures/randPhi.pdf}%
        \vspace{-0.4cm}
        \caption{\textbf{(Left) Evaluating the performance on 60 Atari 2600 games; (Right) Evaluating the performance of \softmoe{} with a random $\boldsymbol{\Phi}$ matrix}. (Left) Even over 60 games, \softmoe{} performs better than the baseline. (Right) Learning $\boldsymbol{\Phi}$ is beneficial over a random phi, indicating that \softmoe{}'s perfomance gains are not only due to the distribution of tokens to experts. Both plots run with Rainbow using the Impala architecture and 8 experts. See \cref{sec:setup} for training details.}
        \label{fig:allsuite}
        \vspace{-0.2cm}
    \end{figure}
\fi

\subsection{Impact of Design Choices}
\label{sec:impactDesign}
\paragraph{Number of experts} \citet{fedus2022switch} argued that the number of experts is the most efficient way to scale models in supervised learning settings. Our results in \cref{fig:topline} demonstrate that while \softmoe{} does benefit from more experts, \moe{} does not. 

\paragraph{Dimensionality of experts} As \cref{fig:topline} confirms, scaling the corresponding layer in the base architectures does not match the performance obtained when using \softmoe{}s (and in fact worsens it). We explored dividing the dimensionality of each expert by the number of experts, effectively bringing the number of parameters on-par with the original base architecture. \cref{fig:rainbowScaling} demonstrates that \softmoe{} maintains its performance, even with much smaller experts. This suggests the observed benefits come largely from the structured sparsity induced by \softmoe{}s, and not necessarily the size of each expert.

\paragraph{Gating and combining} \moe{}s and \softmoe{}s use learned gating mechanisms, albeit different ones: the former uses a top-1 router, selecting an expert for each token, while the latter uses dispatch weights to assign weighted tokens to expert slots. The learned ``combiner'' component takes the output of the MoE modules and combines them to produce a single output (see \cref{fig:moeArchitecture}).  If we use a single expert, the only difference between the base architectures and the MoE ones are then the extra learned parameters from the gating and combination components. \cref{fig:topline} suggests that most of the benefit of MoEs comes from the combination of the gating/combining components with multiple experts. Interestingly, \softmoe{} with a single expert still provides performance gains for Rainbow, suggesting that the learned $\boldsymbol{\Phi}$ matrix (see \cref{sec:moeBackground}) has a beneficial role to play. Indeed, the right panel of \cref{fig:allsuite}, where we replace the learned $\Phi$ with a random one, confirms this.

\begin{figure}[!t]
    \centering
    \includegraphics[width=\linewidth]{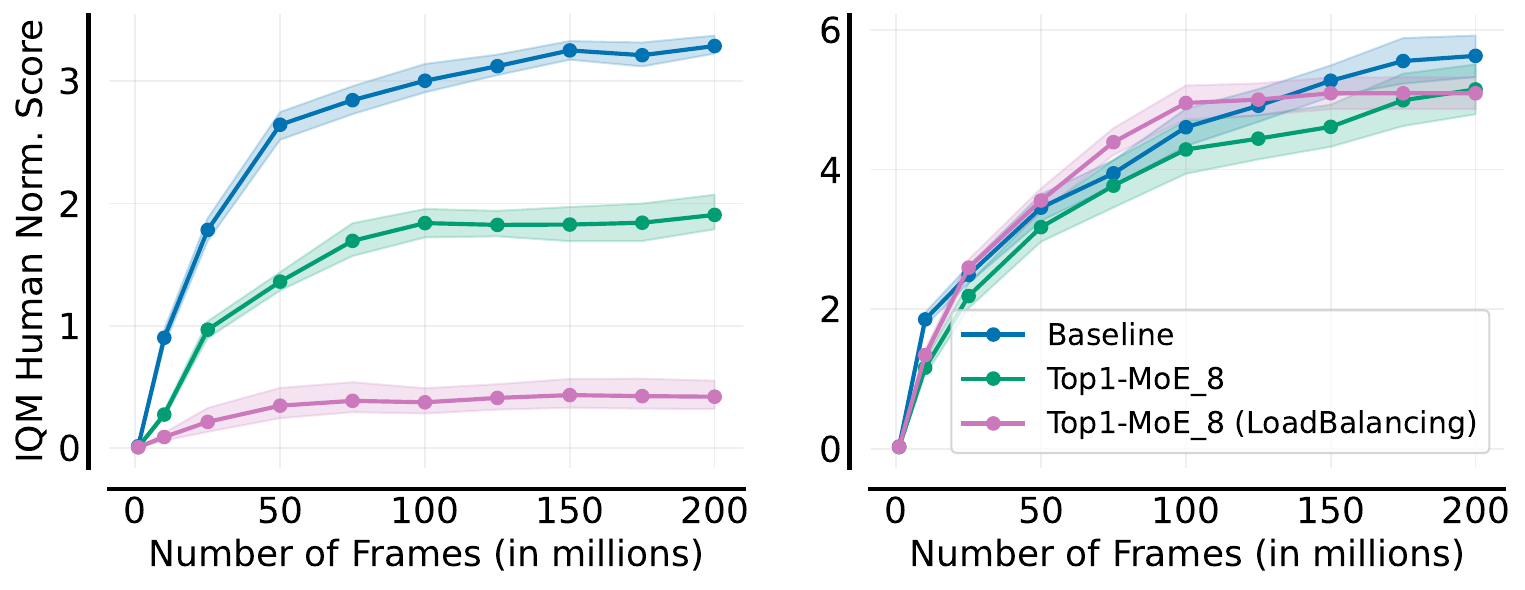}
    \vspace{-0.4cm}
    \caption{{\bf Load-balancing losses from \citet{ruiz2021scaling} are unable to improve the performance of \moe{}} with 8 experts on neither DQN (left) nor Rainbow (right). See \cref{sec:setup} for training details.}
    \label{fig:combinedLoadBalancing}%
\end{figure}

\ifarxiv
    \begin{figure*}[!t]
        \centering
        \includegraphics[width=\linewidth]{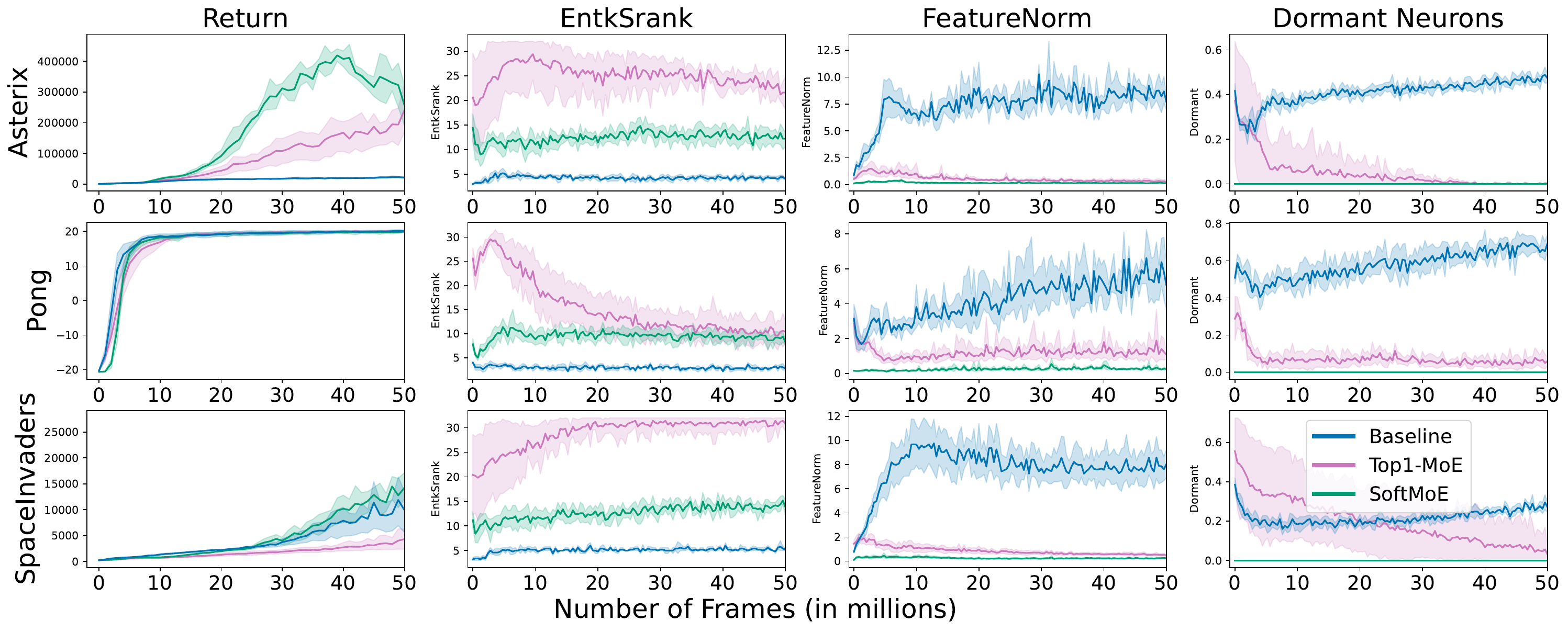}
        \caption{{\bf Additional analyses:} both standard \moe{} and \softmoe{} architectures exhibit similar properties of the hidden representation: both are more robust to dormant neurons and have higher effective rank of both the features and the gradients. We also see that the MoE architectures exhibit less feature norm growth than the baseline. See \cref{sec:setup} for training details.}
        \label{fig:feature-statistics}
    \end{figure*}
\fi

\paragraph{Tokenization} As mentioned above, we focus most of our investigations on \token{PerConv} tokenization, but also explore two others: \token{PerFeat} is essentially a transpose of \token{PerConv}, producing $d$ tokens of dimensionality $h\times w$; \token{PerSamp} uses the entire output of the encoder as a {\em single} token (see \cref{fig:tokenization}). In \cref{fig:rainbowTokenization} we can observe that while \token{PerConv} works best with \softmoe{}, \moe{} seems to benefit more from \token{PerFeat}.

\paragraph{Encoder} Although we have mostly used the encoder from the Impala architecture \citep{espeholt2018impala}, in \cref{fig:CNNEval} we confirm that \softmoe{} still provides benefits when used with the standard CNN architecture from \citet{mnih2015humanlevel}.

\paragraph{Game selection}
To confirm that our findings are not limited to our choice of $20$ games, we ran a study over all the $60$ Atari $2600$ games with $5$ independent seeds, similar to previous works \citep{fedus2020revisiting, ceron2023small}. In the left panel of \cref{fig:allsuite}, we observe that \softmoe{} results in improved performance over the full 60 games in the suite.

\ifarxiv
\else
    \begin{figure*}[!t]
        \centering
        \includegraphics[width=\linewidth]{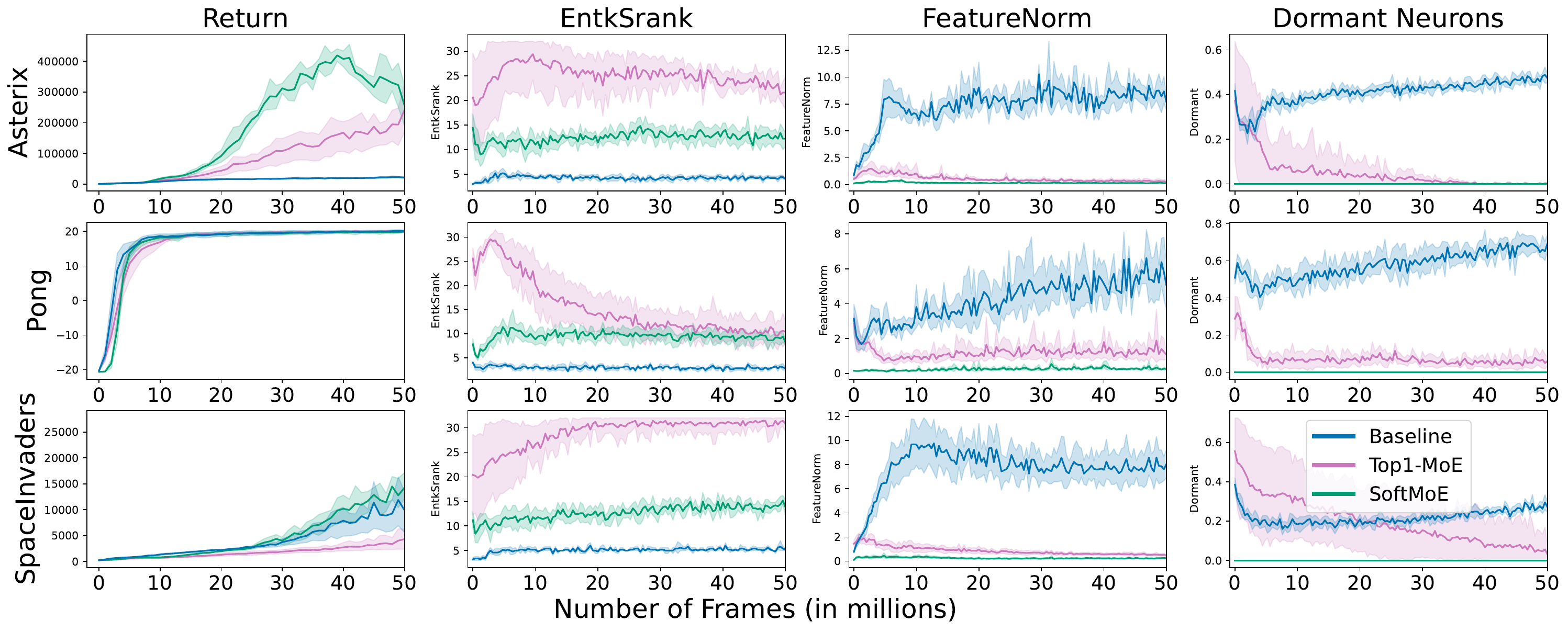}
        \caption{{\bf Additional analyses:} both standard \moe{} and \softmoe{} architectures exhibit similar properties of the hidden representation: both are more robust to dormant neurons and have higher effective rank of both the features and the gradients. We also see that the MoE architectures exhibit less feature norm growth than the baseline. See \cref{sec:setup} for training details.}
        \label{fig:feature-statistics}
    \end{figure*}
\fi

\paragraph{Number of active experts} A crucial difference between the two flavors of MoEs considered here is in the activation of experts: \moe{} activates only {\em one} expert per forward pass (hard-gating), whereas \softmoe{} activates all of them. Hard-gating is known to be a source of training difficulties, and many works have explored adding load-balancing losses \citep{shazeer2017outrageously,ruiz2021scaling,fedus2022switch,mustafa2022multimodal}. To investigate whether \moe{} is underperforming due to improper load balancing, we added the load and importance losses proposed by \citet{ruiz2021scaling} (equation (7) in Appendix 2). \cref{fig:combinedLoadBalancing} suggests the addition of these load-balancing losses is insufficient to boost the performance of \moe{}. While it is possible other losses may result in better performance, these findings suggest that RL agents benefit from having a weighted combination of the tokens, as opposed to hard routing.

\subsection{Additional Analysis}
\label{sec:analyses}
In the previous sections, we have shown that deep RL agents using MoE networks are better able to take advantage of network scaling, but it is not obvious a priori how they produce this effect. While a fine-grained analysis of the effects of MoE modules on network optimization dynamics lies outside the scope of this work, we zoom in on three properties known to correlate with training instability in deep RL agents: the rank of the features \citep{kumar2021implicit}, interference between per-sample gradients \citep{lyle2022learning}, and dormant neurons \citep{sokar2023dormant}. We conduct a deeper investigation into the effect of MoE layers on learning dynamics by studying the Rainbow agent using the Impala architecture, and using 8 experts for the runs with the MoE modules. We track the norm of the features, the rank of the empirical neural tangent kernel (NTK) matrix (i.e. the matrix of dot products between per-transition gradients sampled from the replay buffer), and the number of dormant neurons -- all using a batch size of $32$ -- and visualize these results 
in \cref{fig:feature-statistics}. 

We observe significant differences between the baseline architecture and the architectures that include MoE modules. The MoE architectures both exhibit higher numerical ranks of the empirical NTK matrices than the baseline network, and have negligible dormant neurons and feature norms. These findings suggest that the MoE modules have a stabilizing effect on optimization dynamics, though we refrain from claiming a direct causal link between improvements in these metrics and agent performance. For example, while the rank of the ENTK is higher for the MoE agents, the best-performing agent does not have the highest ENTK rank. The absence of pathological values of these statistics in the MoE networks does however, suggest that whatever the precise causal chain is, the MoE modules have a stabilizing effect on optimization.

\ifarxiv
    \begin{figure}[!t]
        \centering
        \includegraphics[width=0.24\textwidth]{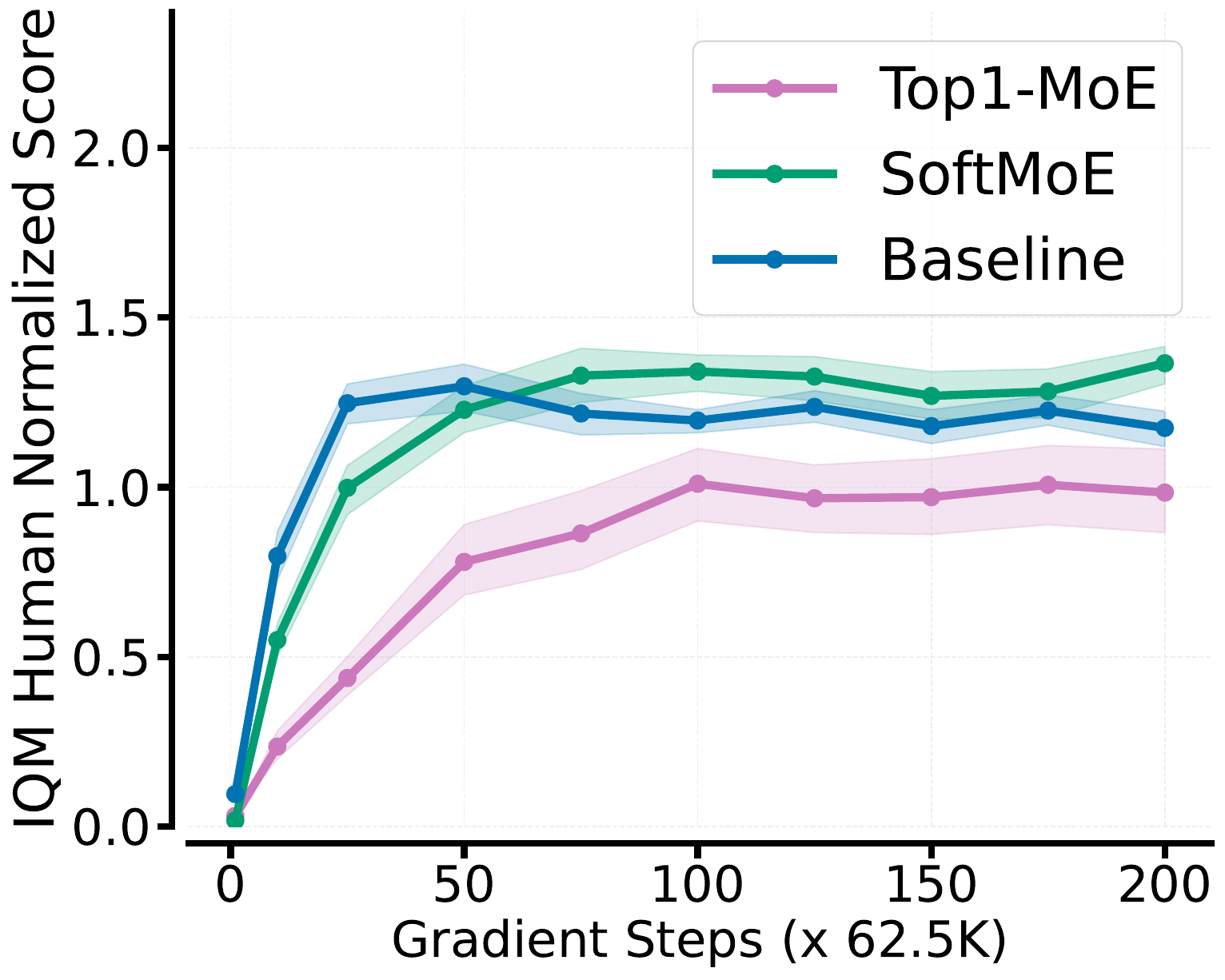}%
        \includegraphics[width=0.24\textwidth]{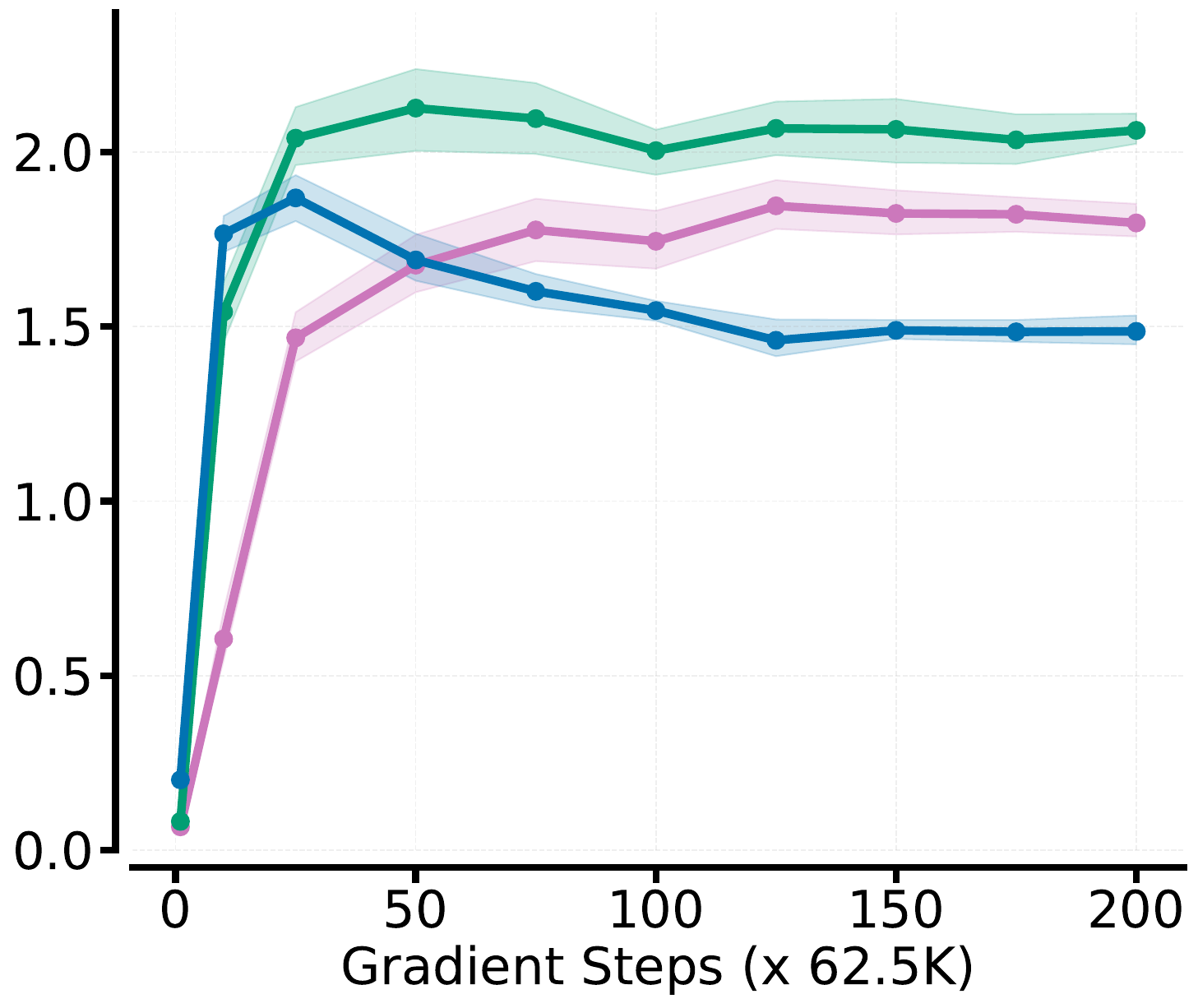}%
        \vspace{-0.4cm}
        \caption{\textbf{Normalized performance across 17 Atari games for CQL (left) and CQL + C51 (right)}, with the ResNet \citep{espeholt2018impala} architecture and 8 experts trained on offline data. \softmoe{} not only remains generally stable with more training, but also attains higher final performance. See \cref{sec:setup} for training details.}
        \label{fig:offline_rlMOEs}
        \vspace{-0.2cm}
    \end{figure}
\fi

\section{Future directions}
\label{sec:futureDirections}
To provide an in-depth investigation into both the resulting performance and probable causes for the gains observed, we focused on evaluating DQN and Rainbow  with \softmoe{} and \moe{} on the standard ALE benchmark. The observed performance gains suggest that these ideas would also be beneficial in other training regimes. In this section we provide strong empirical results in a number of different training regimes, which we hope will serve as indicators for promising future directions of research.

\ifarxiv
    \begin{figure}[!t]
        \centering
        \includegraphics[width=0.24\textwidth]{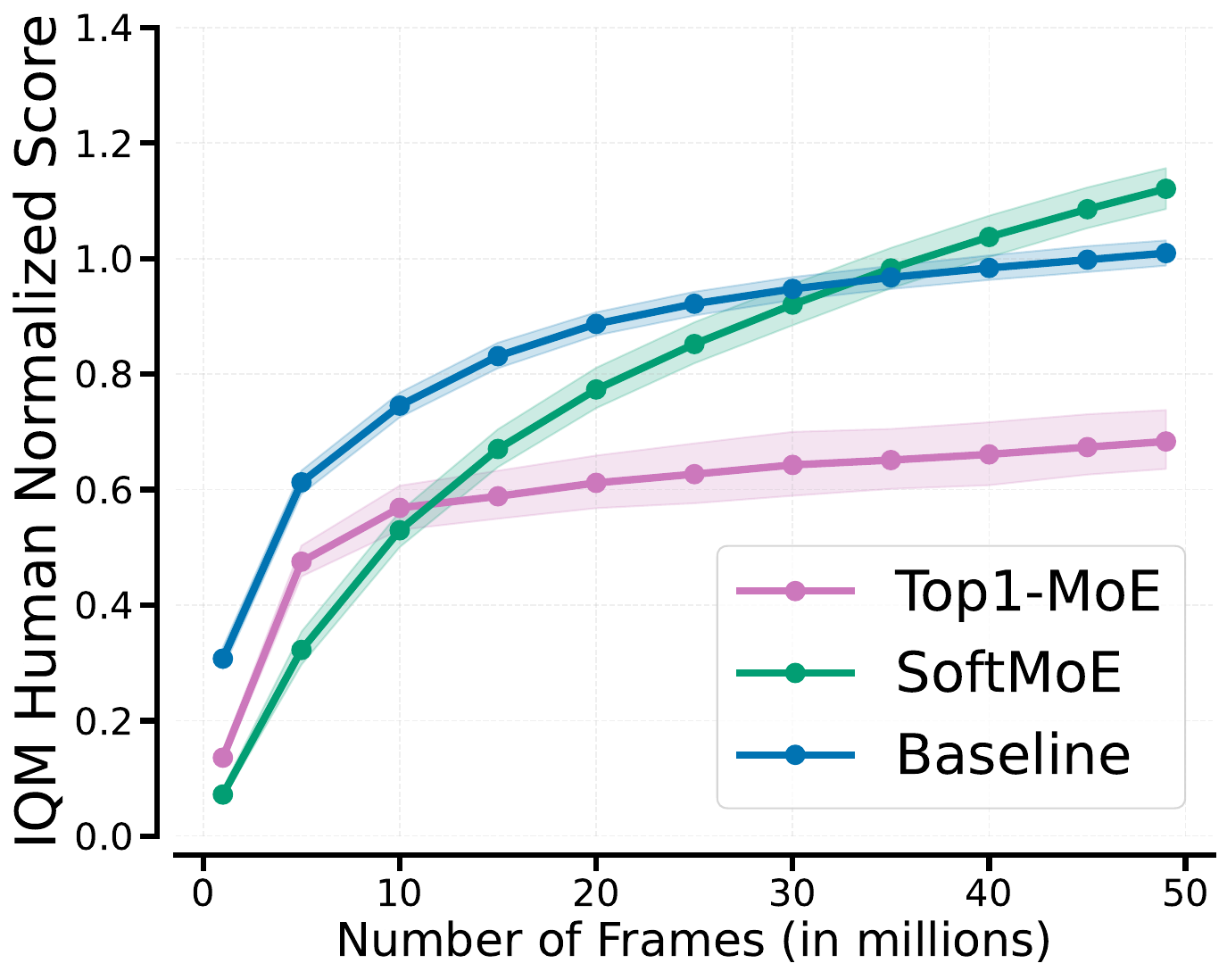}%
        \includegraphics[width=0.24\textwidth]{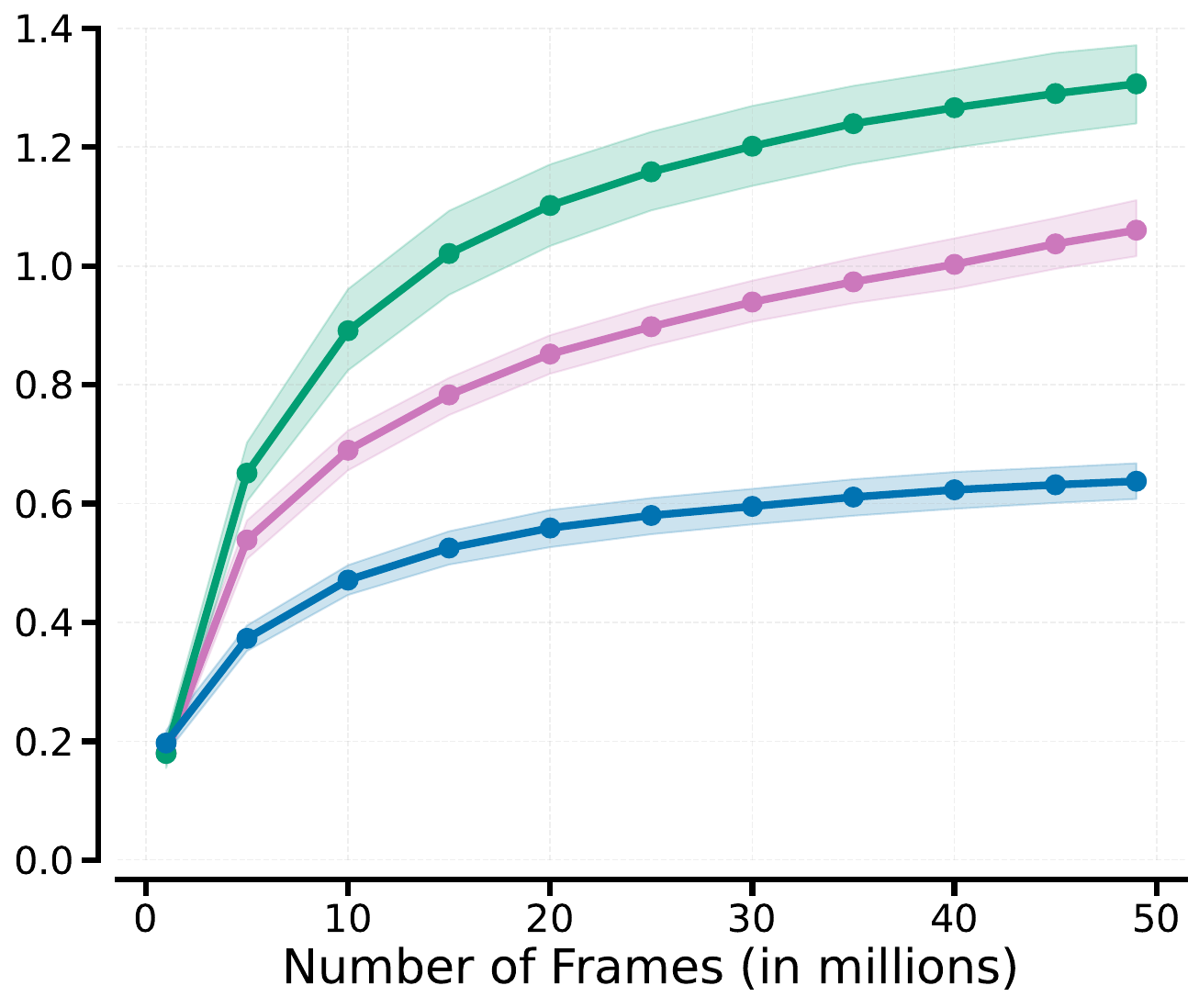}%
        \caption{\textbf{Normalized performance across 26 Atari games for DrQ($\epsilon$) (left) and DER (right)}, with the ResNet architecture \citep{espeholt2018impala} and 8 experts. \softmoe{} not only remains generally stable with more training, but also attains higher final performance. We report interquantile mean performance with error bars indicating 95\% confidence intervals.}
        \label{fig:samplefficiency}
    \end{figure}
\fi

\subsection{Offline RL}\label{sec:offline_rl}
We begin by incorporating MoEs in offline RL, where agents are trained on a fixed dataset without environment interactions. Following prior work \citep{kumar2021dr3}, we train the agents over $17$ games and $5$ seeds for $200$ iterations, where $1$ iteration
corresponds to 62,500 gradient updates. We evaluated on datasets composed of $5\%$, $10\%$, $50\%$  of the samples (drawn randomly) from the set of all environment interactions collected by a DQN agent trained for $200$M steps \citep{agarwal2020optimistic}. In \cref{fig:offline_rlMOEs} we observe that combining \softmoe{} with two modern offline RL algorithms (CQL \citep{kumar2020conservative} and CQL+C51 \citep{kumar2022offline}) attains the best aggregate final performance. \moe{} provides performance improvements when used in conjunction with CQL+C51, but not with CQL. Similar improvements can be observed when using 10\% of the samples (see \cref{append:data_percentage}).

\subsection{Agent Variants For Low-Data Regimes}\label{sec:sample_eff}
DQN and Rainbow were both developed for a training regime where agents can take millions of steps in their environments. \citet{kaiser2020model} introduced the $100$k\footnote{Here, $100$k refers to agent steps, or $400$k environment frames, due to skipping frames in the standard training setup.} benchmark, which evaluates agents on a much smaller data regime ($100$k interactions) on $26$ games.
We evaluate the performance on two popular agents for this regime: DrQ($\epsilon$), an agent based on DQN \citep{yarats2021image,agarwal2021deep}, and DER \citep{van2019use}, an agent based on Rainbow. When trained for 100k environment interactions we saw no real difference between the different variants, suggesting that the benefits of MoEs, at least in our current setup, only arise when trained for a significant amount of interactions. 
However, when trained for 50M steps we do see gains in both agents, in particular with DER (\cref{fig:samplefficiency}). The gains we observe in this setting are consistent with what we have observed so far, given that DrQ($\epsilon$) and DER are based on DQN and Rainbow, respectively.

\vspace{0.2cm}
\subsection{Expert Variants}
\label{sec:arch_exploration}

Our proposed MoE architecture replaces the feed-forward layer after the encoder with an MoE, where the experts consist of a single feed-forward layer (\cref{fig:moeArchitecture}). This is based on what is common practice when adding MoEs to transformer architectures, but is by no means the only way to utilize MoEs.
Here we investigate three variants of using \softmoe{} for DQN and Rainbow with the Impala architecture:
{\bf Big:} Each expert is a full network.The final linear layer is included in DQN (each expert has its own layer), but excluded from Rainbow (so the final linear layer is shared amongst experts).\footnote{This design choice was due to the use of C51 in Rainbow, which makes it non-trivial to maintain`    the ``token-preserving'' property of MoEs.}
{\bf All:} A separate Soft MoE is applied at each layer of the network, excluding the last layer for Rainbow.
{\bf Regular:} the setting used in the rest of this paper.
For {\bf Big} and {\bf All}, the routing is applied at the the input level, and we are using \token{PerSamp} tokenization (see \cref{fig:tokenization}).

\citet{puigcerver2023sparse} propose using $\ell_2$ normalization to each \softmoe{} layer for large tokens, but mention that it makes little difference for smaller tokens. Since the tokens in our experiments above are small (relative to the large models typically using MoEs) we chose not to include this in the experiments run thus far. This may no longer be the case with {\bf Big} and {\bf All}, since we are using \token{PerSamp} tokenization on the full input; for this reason, we investigated the impact of $\ell_2$ normalization when running these results. 

\cref{fig:dqn_bigmoe_iqm} summarizes our results, from which we can draw a number of conclusions. First, these experiments confirm our intuition that {\em not} including regularization in the setup used in the majority of this paper performs best. Second, consistent with \citet{puigcerver2023sparse},  normalization seems to help with {\bf Big} experts. This is particularly noticeable in DQN, where it even surpasses the performance of the Regular setup; the difference between the two is most likely due to the last layer being shared (as in Rainbow) or non-shared (as in DQN). Finally, using separate MoE modules for each layer ({\bf All}) seems to not provide many gains, {\em especially} when coupled with normalization, where the agents are completely unable to learn.

In summary, our results suggest there may be promise in exploring alternative architectures for use with mixtures of experts.

\section{Related Work}
\label{sec:related_work}

\paragraph{Mixture of Experts} MoEs were first proposed by \citet{jacobs1991adaptive} and have recently helped scaling language models up to trillions of parameters thanks to their modular nature, facilitating distributed training, and improved parameter efficiency at inference~\citep{lepikhin2020gshard,fedus2022switch}.
MoEs are also widely studied in computer vision \cite{wang2020deep,yang2019condconv,abbas2020biased,pavlitskaya2020using}, where they enable scaling vision transformers to billions of parameters while reducing the inference computation cost by half \citep{riquelme2021scaling}, help with vision-based continual learning \citep{lee2019neural}, and multi-task problems~\citep{fan2022m3vit}.   
MoEs also help performance in transfer- and multi-task learning settings, e.g. by specializing experts to sub-problems~\citep{puigcerver2020scalable, chen2023mod, ye2023taskexpert} or by addressing statistical performance issues of routers~\citep{hazimeh2021dselect}.
There have been few works exploring MoEs in RL for single \citep{ren21probabilistic, akrour22continuous} and multi-task learning \citep{hendawy2024multitask}. However their definition and usage of ``mixtures-of-experts'' is somewhat different than ours, focusing more on orthogonal, probabilistic, and interpretable MoEs.

\ifarxiv
\else
    \begin{figure}[!t]
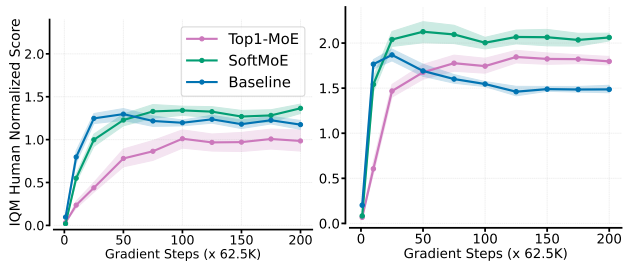

        \centering
        \includegraphics[width=0.24\textwidth]{figures/MOEs_offlineCQL_experts_8CORRCOLOR2.pdf}%
        \includegraphics[width=0.24\textwidth]{figures/MOEs_offlineCQL+C51_experts_8CORRCOLOR2.pdf}%
        \vspace{-0.4cm}
        \caption{\textbf{Normalized performance across $17$ Atari games for CQL (left) and CQL + C51 (right)}, with the ResNet \citep{espeholt2018impala} architecture and $8$ experts trained on offline data. \softmoe{} not only remains generally stable with more training, but also attains higher final performance. See \cref{sec:setup} for training details.}
        \label{fig:offline_rlMOEs}
        \vspace{-0.2cm}
    \end{figure}
\fi

\paragraph{Parameter Scalability and Efficiency in Deep RL}
Lack of parameter scalability in deep RL can be partly explained by a lack of parameter efficiency. Parameter count cannot be scaled efficiently if those parameters are not used effectively. Recent work shows that networks in RL under-utilize their parameters. \citet{sokar2023dormant} demonstrates that networks suffer from an increasing number of inactive neurons throughout online training. Similar behavior is observed by \citet{gulcehre2022empirical} in offline RL. \citet{arnob2021single} shows that 95\% of the network parameters can be pruned at initialization in offline RL without loss in performance. Numerous works demonstrate that periodic resetting of the network weights improves performance \citep{nikishin22primacy,dohare2021continual,sokar2023dormant,d'oro2022sampleefficient,igl2020transient,schwarzer23bbf}. 

Another line of research demonstrates that RL networks can be trained with a high sparsity level ($\sim$90\%) without loss in performance \cite{tan2022rlx2,sokar2022dynamic,graesser2022state, obando2024deep}. These observations call for techniques to better utilize the network parameters in RL training, such as using MoEs, which we show decreases dormant neurons drastically over multiple tasks and architectures. To enable scaling networks in deep RL, prior works focus on algorithmic methods \citep{farebrother2022proto, taiga2022investigating, schwarzer23bbf, farebrother24classification}. In contrast, we focus on alternative {\em network topologies} to enable robustness towards scaling.

\section{Discussion and Conclusion}
\label{sec:Discussion and Conclusion}

\ifarxiv
\else
    \begin{figure}[!t]
        \centering
        \includegraphics[width=0.24\textwidth]{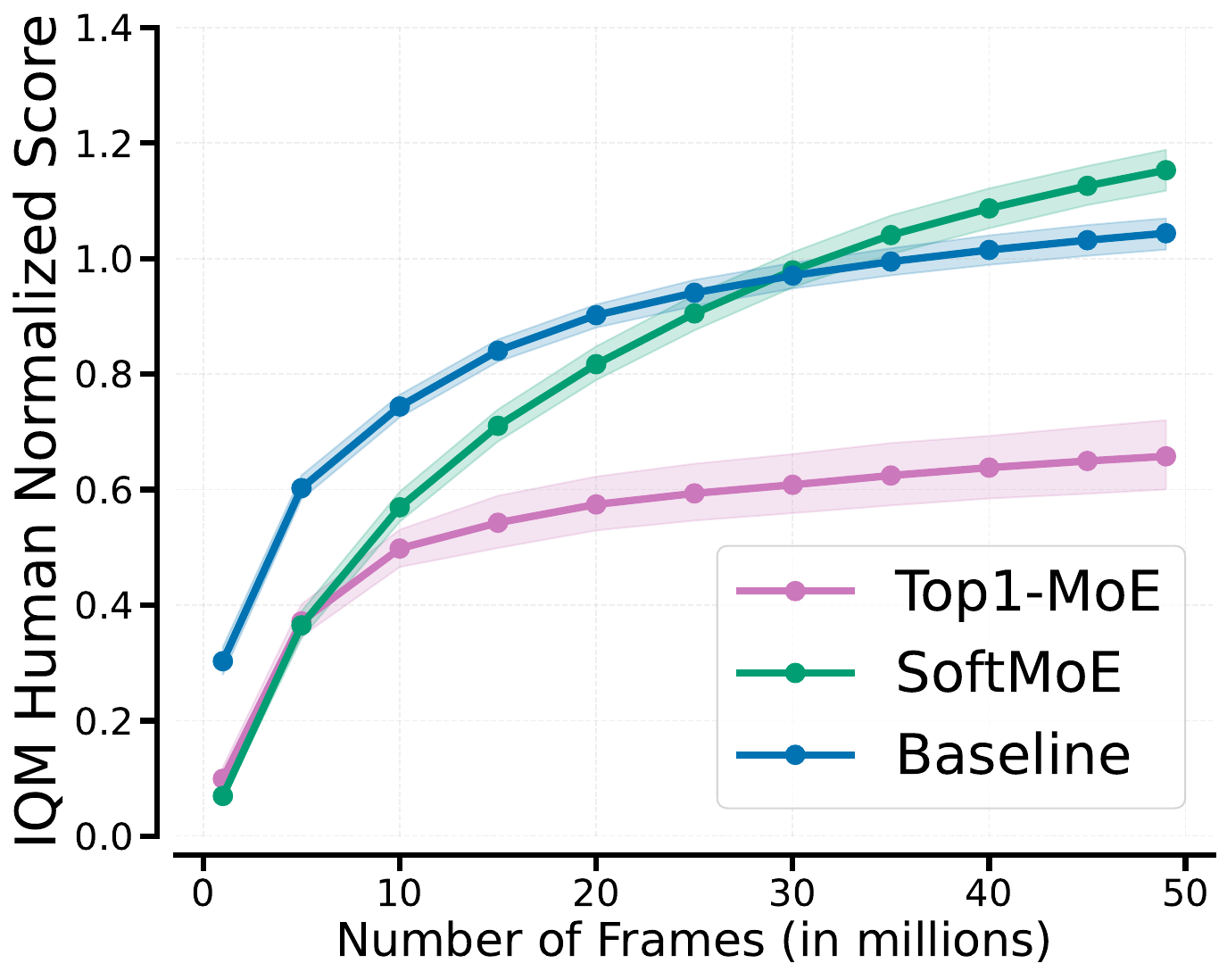}%
         \includegraphics[width=0.24\textwidth]{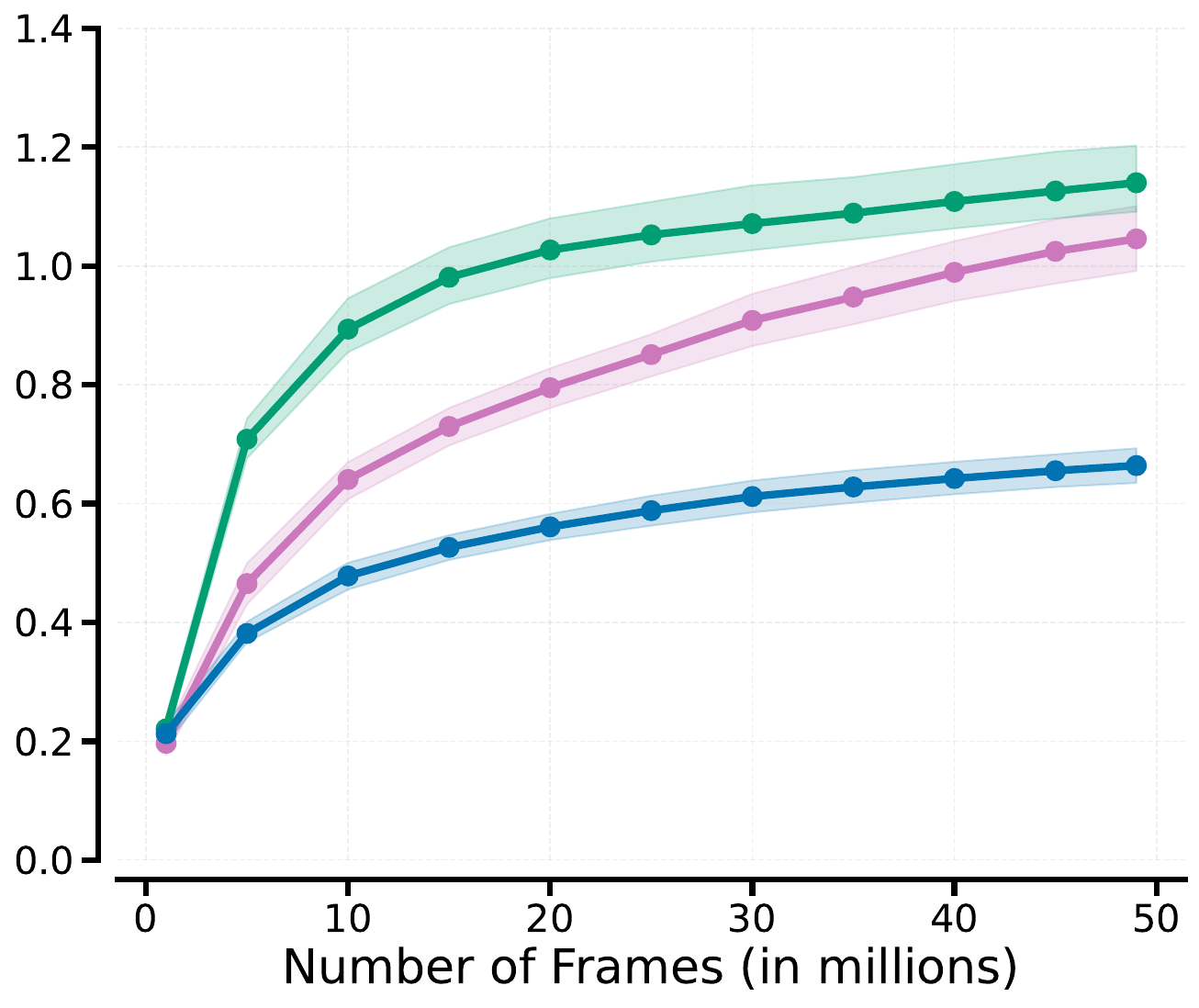}%
        \vspace{-0.4cm}
        \caption{\textbf{Normalized performance across $26$ Atari games for DrQ($\epsilon$) (left) and DER (right)}, with the ResNet architecture \citep{espeholt2018impala} and $8$ experts (see \autoref{fig:samplefficiency_4experts}, for $4$ experts). \softmoe{} not only remains generally stable with more training, but also attains higher final performance. We report interquantile mean performance with error bars indicating $95\%$ confidence intervals.}
        \label{fig:samplefficiency}
    \end{figure}
\fi

As RL continues to be used for increasingly complex tasks, we will likely require larger networks. As recent research has shown (and which our results confirm), na\"{i}vely scaling up network parameters does not result in improved performance. Our work shows empirically that MoEs have a beneficial effect on the performance of value-based agents across a diverse set of training regimes.

Mixtures of Experts induce a form of {\em structured sparsity} in neural networks, prompting the question of whether the benefits we observe are simply a consequence of this sparsity rather than the MoE modules themselves. Our results suggest that it is likely a combination of both: \cref{fig:topline} demonstrates that in Rainbow, adding an MoE module with a {\em single} expert yields statistically significant performance improvements, while \cref{fig:rainbowScaling} demonstrates that one can scale down expert dimensionality without sacrificing performance. The right panel of \cref{fig:allsuite} further confirms the necessity of the extra parameters in \softmoe{} modules.

Recent findings in the literature demonstrate that while RL networks have a natural tendency towards neuron-level sparsity which can hurt performance \citep{sokar2023dormant}, they can benefit greatly from explicit parameter-level sparsity \citep{graesser2022state}. When taken in combination with our findings, they suggest that there is still much room for exploration, and understanding, of the role sparsity can play in training deep RL networks, especially for parameter scalability.

Even in the narrow setting we have focused on (replacing the penultimate layer of value-based agents with MoEs for off-policy learning in single-task settings), there are many open questions that can further increase the benefits of MoEs: different values of $k$ for \moe{}s, different tokenization choices, using different learning rates (and perhaps optimizers) for routers, among others. Of course, expanding beyond the ALE could provide more comprehensive results and insights, potentially at a fraction of the computational expense \citep{ceron2021revisiting}.

\ifarxiv
\else
    \begin{figure}[!t]
        \centering
        \includegraphics[width=0.241\textwidth]{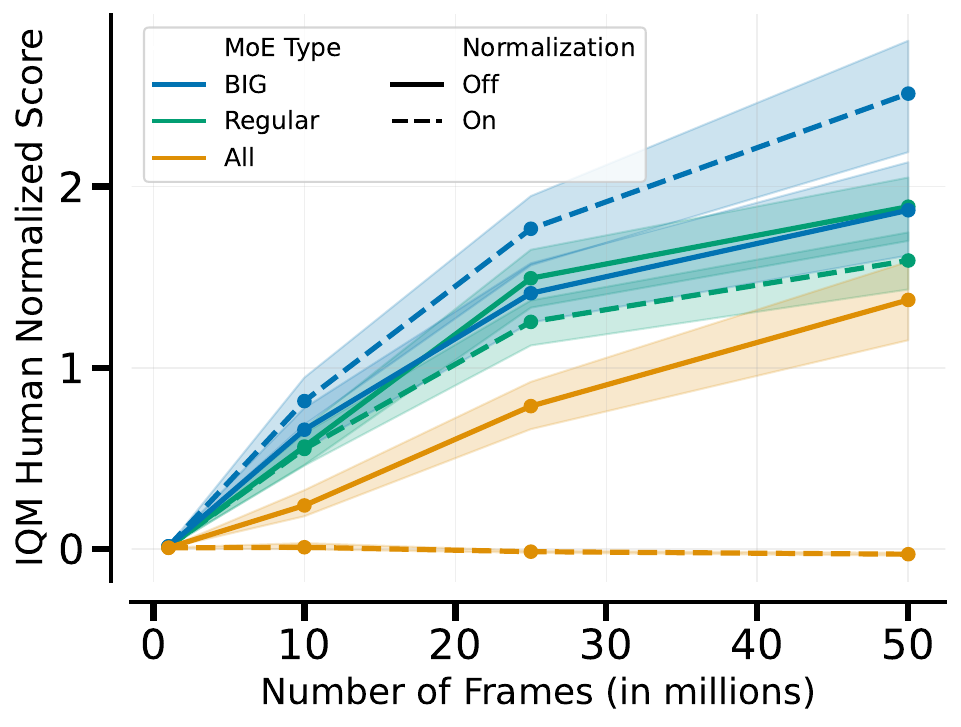}%
        \includegraphics[width=0.241\textwidth]{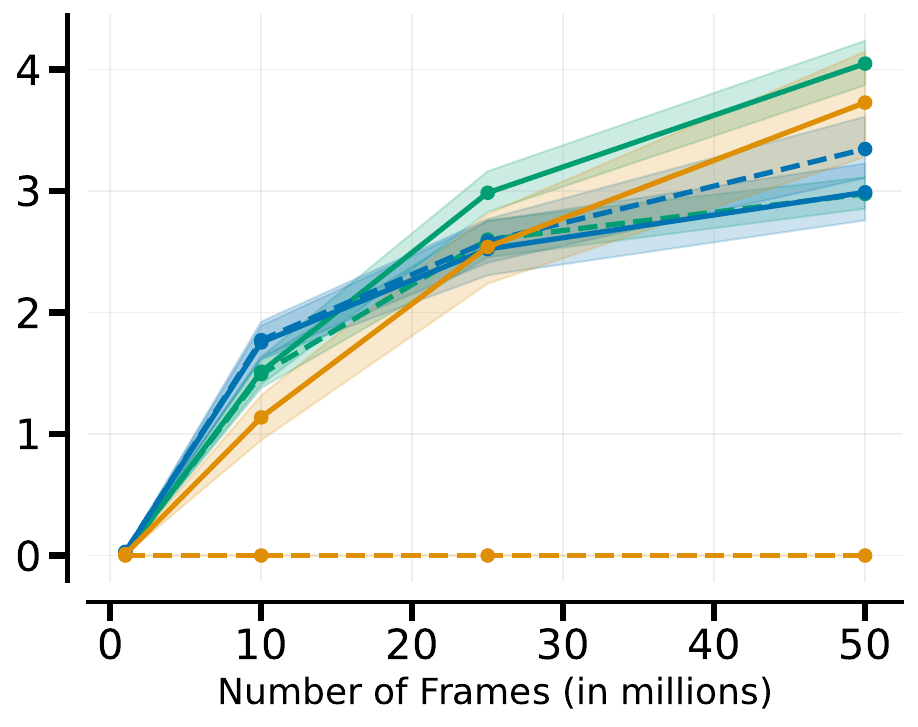}%
        
        \caption{\textbf{Normalized performance over $20$ games for expert variants} with DQN (left) and Rainbow (right), also investigating the use of the normalization of \citet{puigcerver2023sparse}. We report interquantile mean performance with shaded areas indicating 95\% confidence intervals.}
        \label{fig:dqn_bigmoe_iqm}
    \end{figure}
\fi

The results presented in \cref{sec:futureDirections} suggest MoEs can play a more generally advantageous role in training deep RL agents. More broadly, our findings confirm the impact architectural design choices can have on the ultimate performance of RL agents. We hope our findings encourage more researchers to further investigate this -- still relatively unexplored -- research direction.

\ifarxiv
\else
    \section*{Acknowledgements}
The authors would like to thank Gheorghe Comanici, Owen He, Alex Muzio, Adrien Ali Ta\"{i}ga, Rishabh Agarwal, Hugo Larochelle, Ayoub Echchahed, and the rest of the Google DeepMind Montreal team for valuable discussions during the preparation of this work; Gheorghe deserves a special mention for providing us valuable feedback on an early draft of the paper. We thank the anonymous reviewers for their valuable help in improving our manuscript. We would also like to thank the Python community \citep{van1995python, 4160250} for developing tools that enabled this work, including NumPy \cite{harris2020array}, Matplotlib \cite{hunter2007matplotlib}, Jupyter \cite{2016ppap}, Pandas \cite{McKinney2013Python} and JAX \cite{bradbury2018jax}.
    \section*{Impact statement}
This paper presents work whose goal is to advance the field of Machine Learning, and reinforcement learning in particular. There are many potential societal consequences of our work, none which we feel must be specifically highlighted here.
\fi

\bibliography{references}
\ifarxiv
\else
    \bibliographystyle{icml2024}
\fi

\newpage
\appendix
\onecolumn
\section{Experimental details}
\label{sec:atari_2600}

Unless otherwise specified, in all experiments below we report the interquantile mean after $40$ million environment steps; error bars indicate $95\%$ stratified bootstrap confidence intervals \citep{agarwal2021deep}. Most of our experiments were run with $20$ games from the ALE suite \citep{Bellemare_2013}, as suggested by \citet{fedus2020revisiting}. However, for the Atari $100k$ agents (\autoref{sec:sample_eff}), we used the standard set of $26$ games \citep{kaiser2020model} to be consistent with the benchmark. Finally, we also ran some experiments with the full set of $60$ games. The specific games are detailed below.

\textbf{20 game subset:} AirRaid, Asterix, Asteroids, Bowling, Breakout, DemonAttack, Freeway, Gravitar, Jamesbond, MontezumaRevenge, MsPacman, Pong, PrivateEye, Qbert, Seaquest, SpaceInvaders, Venture,         WizardOfWor, YarsRevenge, Zaxxon.

\textbf{26 game subset:} Alien, Amidar, Assault, Asterix, BankHeist, BattleZone, Boxing, Breakout, ChopperCommand, CrazyClimber, DemonAttack, Freeway, Frostbite, Gopher, Hero, Jamesbond, Kangaroo, Krull, KungFuMaster, MsPacman, Pong, PrivateEye, Qbert, RoadRunner, Seaquest, UpNDown.

\textbf{60 game set:} The 26 games above in addition to: AirRaid, Asteroids, Atlantis, BeamRider, Berzerk, Bowling, Carnival, Centipede, DoubleDunk, ElevatorAction, Enduro, FishingDerby, Gravitar, IceHockey, JourneyEscape, MontezumaRevenge, NameThisGame, Phoenix, Pitfall, Pooyan, Riverraid, Robotank, Skiing, Solaris, SpaceInvaders, StarGunner, Tennis, TimePilot, Tutankham, Venture, VideoPinball, WizardOfWor, YarsRevenge, Zaxxon.

{
\section{Hyper-parameters list}
\label{sec:list_hyperparameters}

Default hyper-parameter settings for DER \citep{van2019use} and DrQ($\epsilon$) \citep{kaiser2020model,agarwal2021deep} agents. \autoref{tbl:defaultvalues} shows the default values for each hyper-parameter across all the Atari games.

\begin{table}[!h]
 \centering
  \caption{Default hyper-parameters setting for DER and DrQ($\epsilon$) agents.}
  \label{tbl:defaultvalues}
 \begin{tabular}{@{} ccc @{}}
    \toprule
    & \multicolumn{2}{c}{Atari}\\
    \cmidrule(lr){2-3}
  Hyper-parameter &  DER & DrQ($\epsilon$) \\
    \midrule
     Adam's($\epsilon$) & 0.00015 & 0.00015\\
     Adam's learning rate & 0.0001 & 0.0001 \\
     Batch Size & 32 & 32\\
     Conv. Activation Function & ReLU & ReLU \\
     Convolutional Width & 1& 1\\
     Dense Activation Function & ReLU & ReLU\\
     Dense Width & 512 & 512 \\
     Normalization & None & None \\
     Discount Factor & 0.99 & 0.99 \\
     Exploration $\epsilon$ & 0.01 & 0.01\\
     Exploration $\epsilon$ decay & 2000 & 5000\\
     Minimum Replay History & 1600 & 1600\\
     Number of Atoms & 51 & 0 \\
     Number of Convolutional Layers & 3 & 3\\
     Number of Dense Layers & 2 & 2\\
     Replay Capacity & 1000000 & 1000000 \\
     Reward Clipping & True & True \\
     Update Horizon & 10 & 10 \\
     Update Period & 1& 1\\
     Weight Decay & 0 & 0\\
     Sticky Actions & False & False \\
     \bottomrule
  \end{tabular}
\end{table}

\newpage
Default hyper-parameter settings for DQN \citep{mnih2015humanlevel} and Rainbow \citep{Hessel2018RainbowCI} agents. \autoref{tbl:defaultvalues_40M} shows the default values for each hyper-parameter across all the Atari games.

\begin{table}[!h]
 \centering
  \caption{Default hyper-parameters setting for DQN and Rainbow agents.}
  \label{tbl:defaultvalues_40M}
 \begin{tabular}{@{} ccc @{}}
    \toprule
    & \multicolumn{2}{c}{Atari}\\
    \cmidrule(lr){2-3}
  Hyper-parameter &  DQN & Rainbow \\
    \midrule
     Adam's ($\epsilon$) & 1.5e-4 & 1.5e-4\\
     Adam's learning rate &  6.25e-5 & 6.25e-5 \\
     Batch Size & 32 & 32\\
     Conv. Activation Function & ReLU & ReLU \\
     Convolutional Width & 1 & 1\\
     Dense Activation Function & ReLU & ReLU \\
     Dense Width & 512 & 512 \\
     Normalization & None & None \\
     Discount Factor & 0.99 & 0.99 \\
     Exploration $\epsilon$ & 0.01 & 0.01\\
     Exploration $\epsilon$ decay & 250000 & 250000\\
     Minimum Replay History & 20000 & 20000 \\
     Number of Atoms & 0 & 51 \\
     Number of Convolutional Layers & 3 & 3 \\
     Number of Dense Layers & 2 & 2\\
     Replay Capacity & 1000000 & 1000000  \\
     Reward Clipping & True & True \\
     Update Horizon & 1 & 3\\
     Update Period & 4 & 4   \\
     Weight Decay & 0 & 0 \\
     Sticky Actions & True & True\\
     \bottomrule
  \end{tabular}
\end{table}

\newpage
Default hyper-parameter settings for CQL \citep{kumar2020conservative} and CQL+C51 \citep{kumar2022offline} offline agents. \autoref{tbl:defaultvalues_offline} shows the default values for each hyper-parameter across all the Atari games.

\begin{table}[!h]
 \centering
  \caption{Default hyper-parameters setting for CQL and CQL+C51 agents.}
  \label{tbl:defaultvalues_offline}
 \begin{tabular}{@{} ccc @{}}
    \toprule
    & \multicolumn{2}{c}{Atari}\\
    \cmidrule(lr){2-3}
  Hyper-parameter &  CQL & CQL+C51\\
    \midrule
     Adam's($\epsilon$) & 0.0003125 & 0.00015\\
     Batch Size & 32 & 32\\
     Conv. Activation Function & ReLU & ReLU \\
     Convolutional Width & 1& 1\\
     Dense Activation Function & ReLU & ReLU\\
     Normalization & None & None \\
     Dense Width & 512 & 512 \\
     Discount Factor & 0.99 & 0.99 \\
     Learning Rate & 0.00005 & 0.0000625 \\
     Number of Atoms & 0 & 51 \\
     Number of Convolutional Layers & 3 & 3\\
     Number of Dense Layers & 2 & 2\\
     Fixed Replay Capacity & 2,500,000 steps & 2,500,000 steps \\
     Reward Clipping & True & True \\
     Update Horizon & 1 & 3 \\
     Update Period & 1 & 1\\
     Weight Decay & 0 & 0\\
     Replay Scheme & Uniform & Uniform \\
     Dueling & False & True \\
     Double DQN & False & True \\
     CQL coef & 0.1 & 0.1\\
     \bottomrule
  \end{tabular}
\end{table}

Default hyper-parameter settings for CNN architecture \citep{mnih2015humanlevel} and Impala-based ResNet \citep{espeholt2018impala} \autoref{tbl:defaultvalues_networ} shows the default values for each hyper-parameter across all the Atari games.

\begin{table}[!h]
 \centering
  \caption{Default hyper-parameters for neural networks.}
  \label{tbl:defaultvalues_networ}
 \begin{tabular}{@{} ccc @{}}
    \toprule
    & \multicolumn{2}{c}{Atari}\\
    \cmidrule(lr){2-3}
  Hyper-parameter &  CNN architecture \citep{mnih2015humanlevel} & Impala-based ResNet \citep{espeholt2018impala}\\
    \midrule
    Observation down-sampling & (84, 84) & (84, 84) \\
    Frames stacked &4  &4 \\
     Q-network (channels) & 32, 64, 64 & 32, 64, 64\\
     Q-network (filter size) & 8 x 8, 4 x 4, 3 x 3 & 8 x 8, 4 x 4, 3 x 3\\
     Q-network (stride)  & 4, 2, 1 & 4, 2, 1 \\
     Num blocks & - & 2 \\
     Use max pooling & False & True \\
     Skip connections & False & True \\
     Hardware & Tesla P100 GPU & Tesla P100 GPU\\
     \bottomrule
  \end{tabular}
\end{table}
}

\newpage
\section{Extra results}

\begin{figure}[!h]
    \centering
    \includegraphics[width=\textwidth]{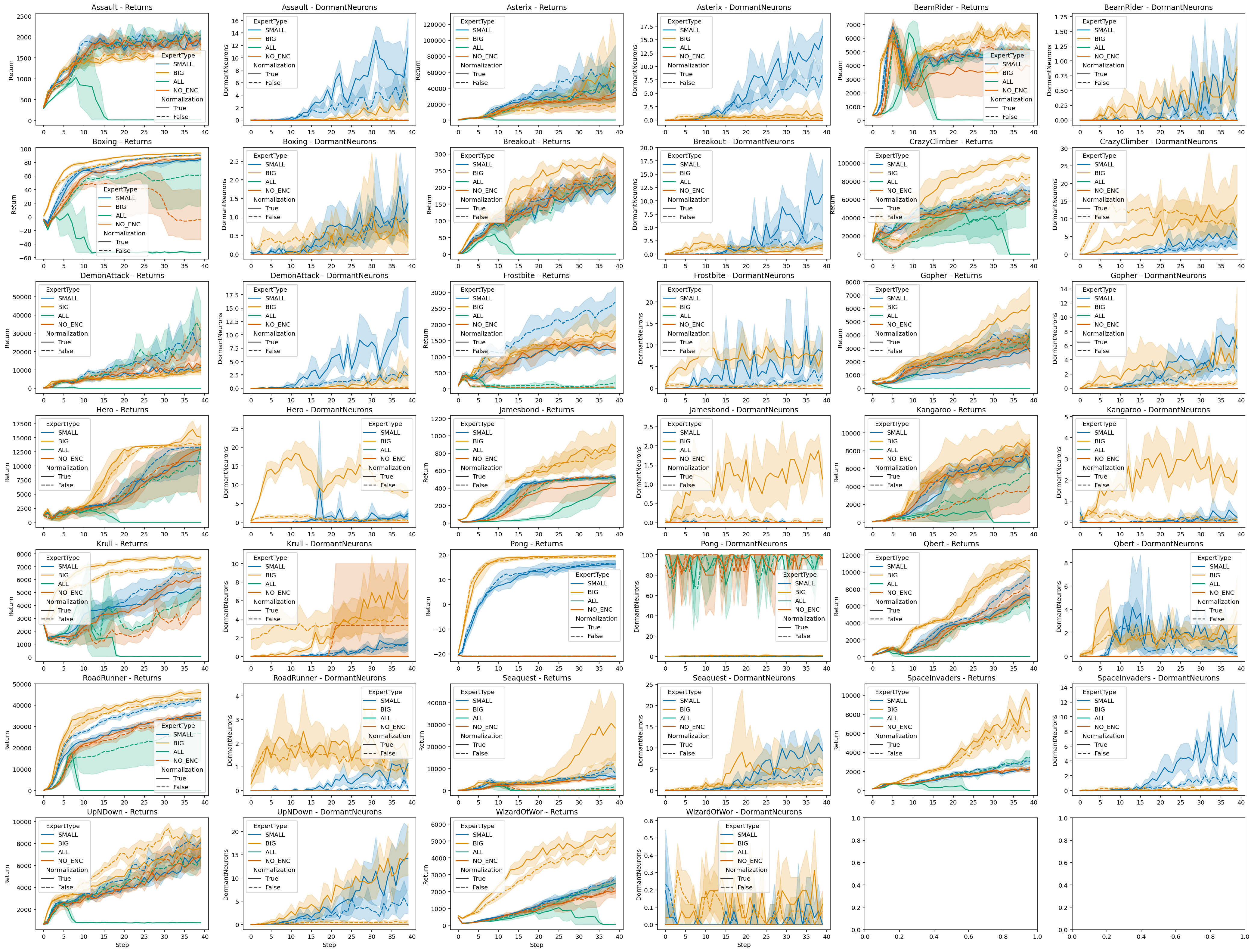}
    \caption{Results for architectural ablations as described in Section \ref{sec:arch_exploration} on DQN. Additionally, we investigate the effect of the normalization that was proposed in the original Soft MoE paper.}
    \label{fig:dqn_bigmoe_20_games}
\end{figure}

\begin{figure}[!h]
    \centering
    \includegraphics[width=\textwidth]{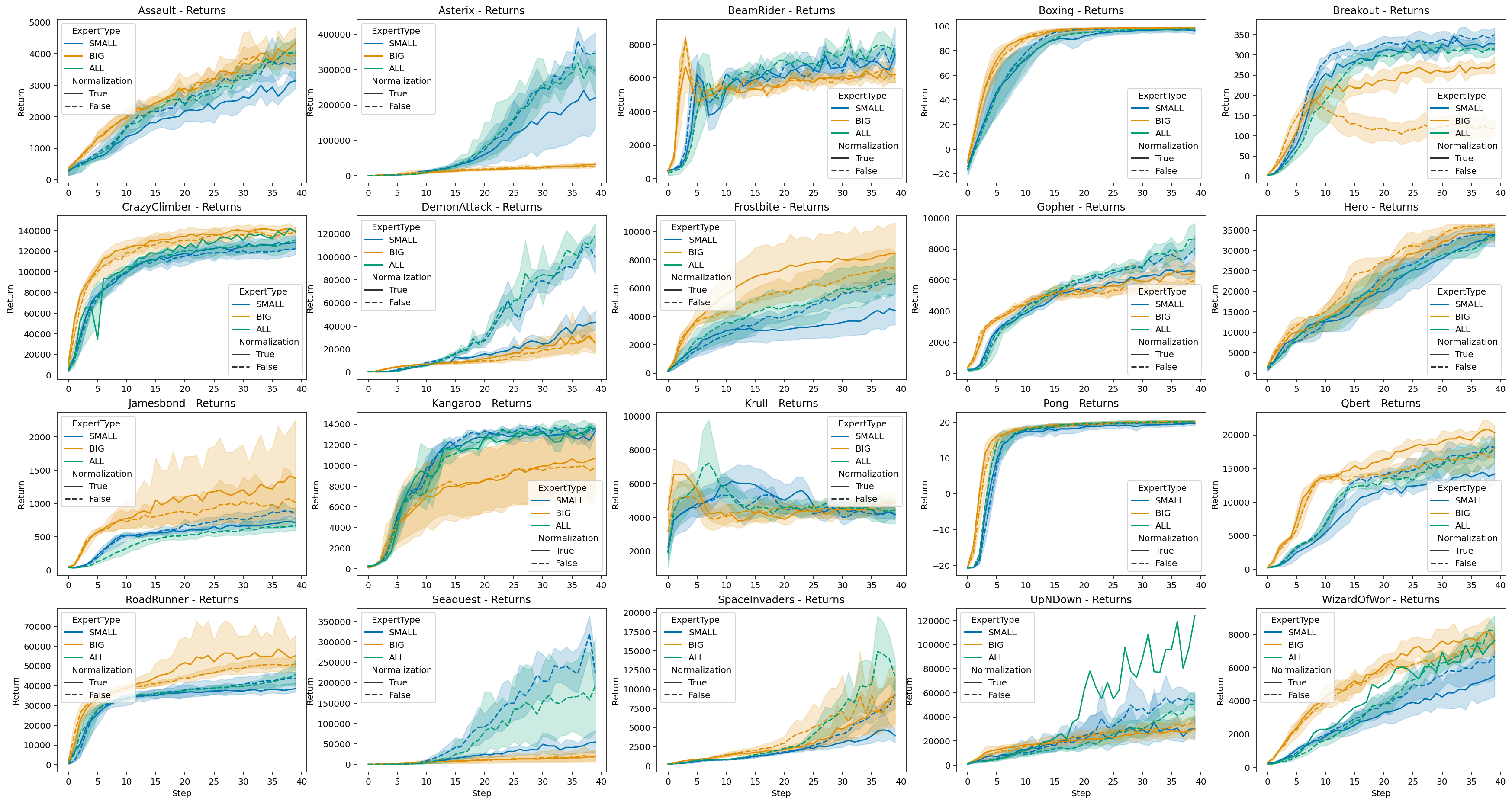}
    \caption{Results for architectural exploration as described in Section \ref{sec:arch_exploration} on Rainbow. Additionally, we investigate the effect of the normalization that was proposed in the original Soft MoE paper.}
    \label{fig:rainbow_bigmoe_20_games}
\end{figure}


\begin{figure}[!h]
    \centering
    \includegraphics[width=0.49\textwidth]{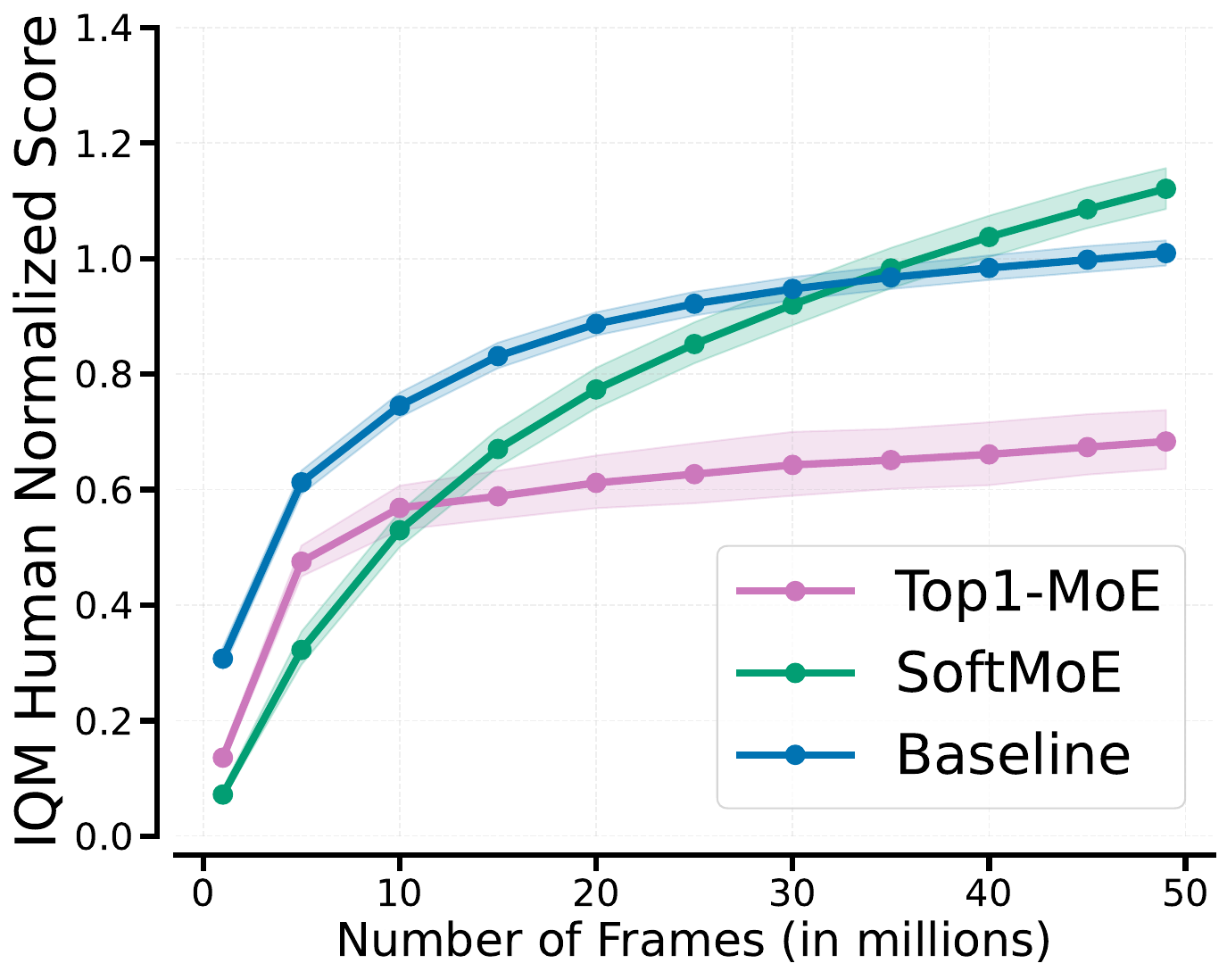}%
    \includegraphics[width=0.49\textwidth]{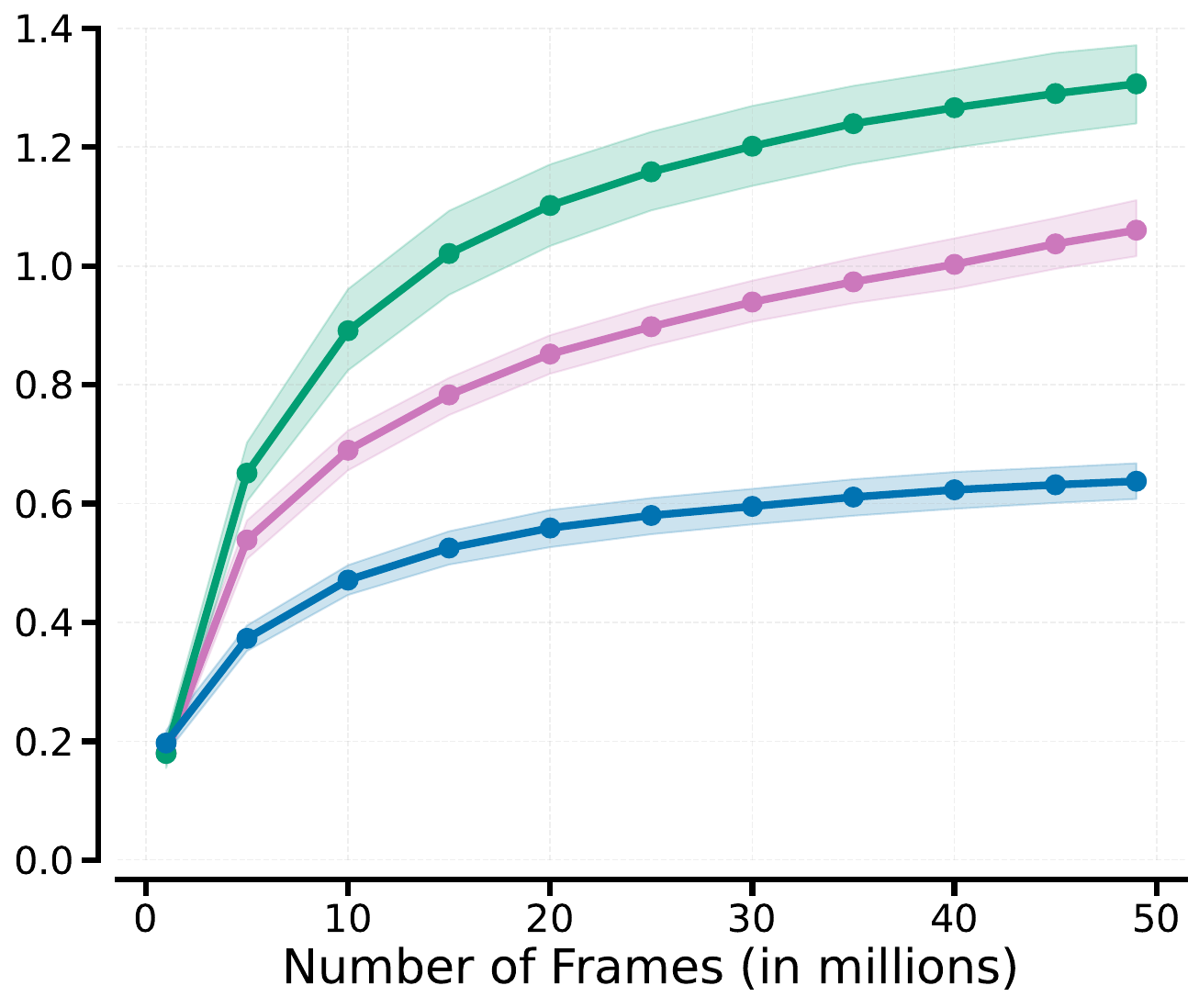}%
    \caption{\textbf{Normalized performance across $26$ Atari games for DrQ($\epsilon$) (left) and DER (right)}, with the ResNet architecture \citep{espeholt2018impala} and $4$ experts. \softmoe{} not only remains generally stable with more training, but also attains higher final performance. We report interquantile mean performance with error bars indicating $95\%$ confidence intervals.}
    \label{fig:samplefficiency_4experts}
    \vspace{-0.2cm}
\end{figure}
    
\begin{figure}[!h]
    \centering
    \includegraphics[width=0.49\textwidth]{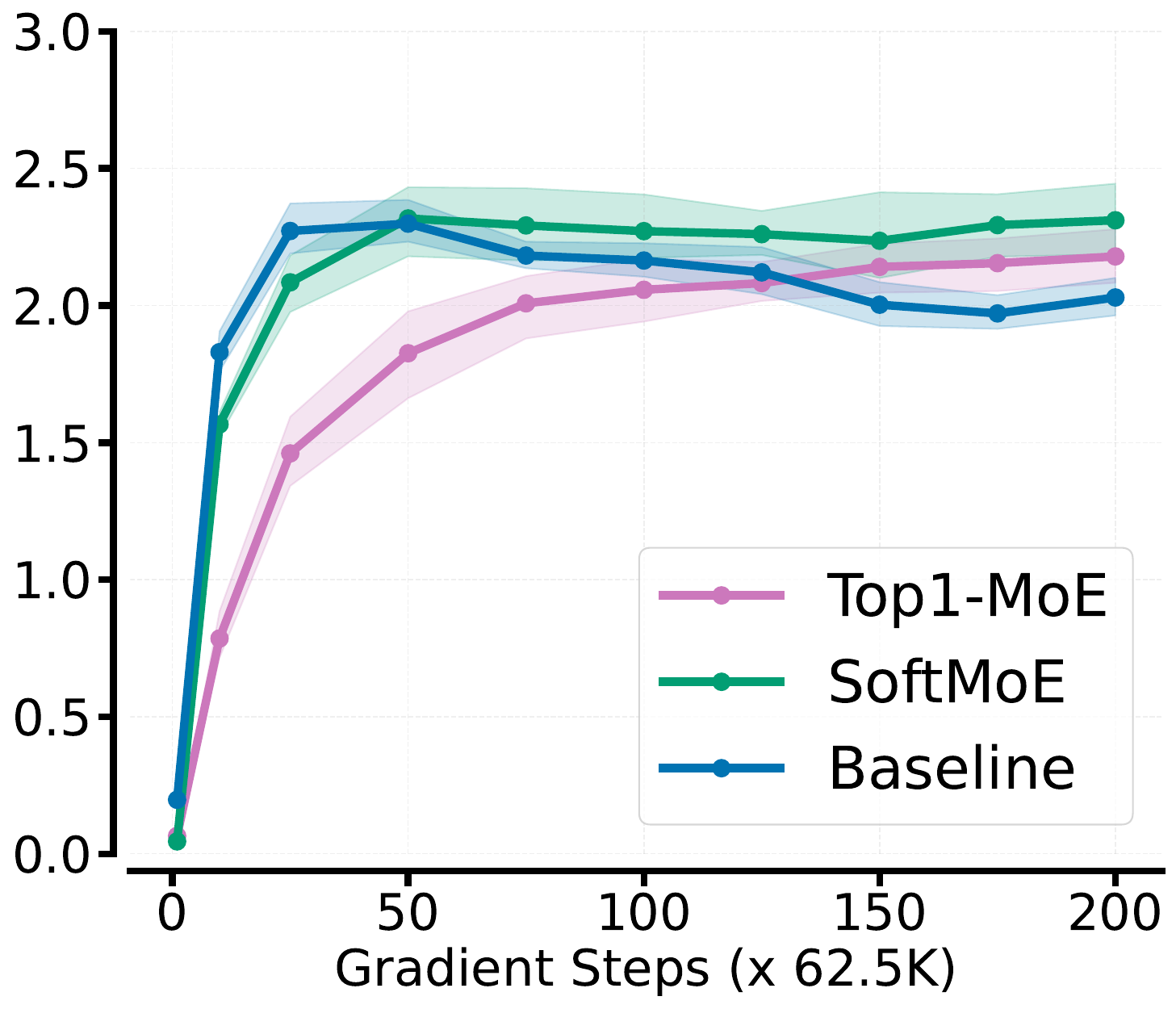}
    \includegraphics[width=0.49\textwidth]{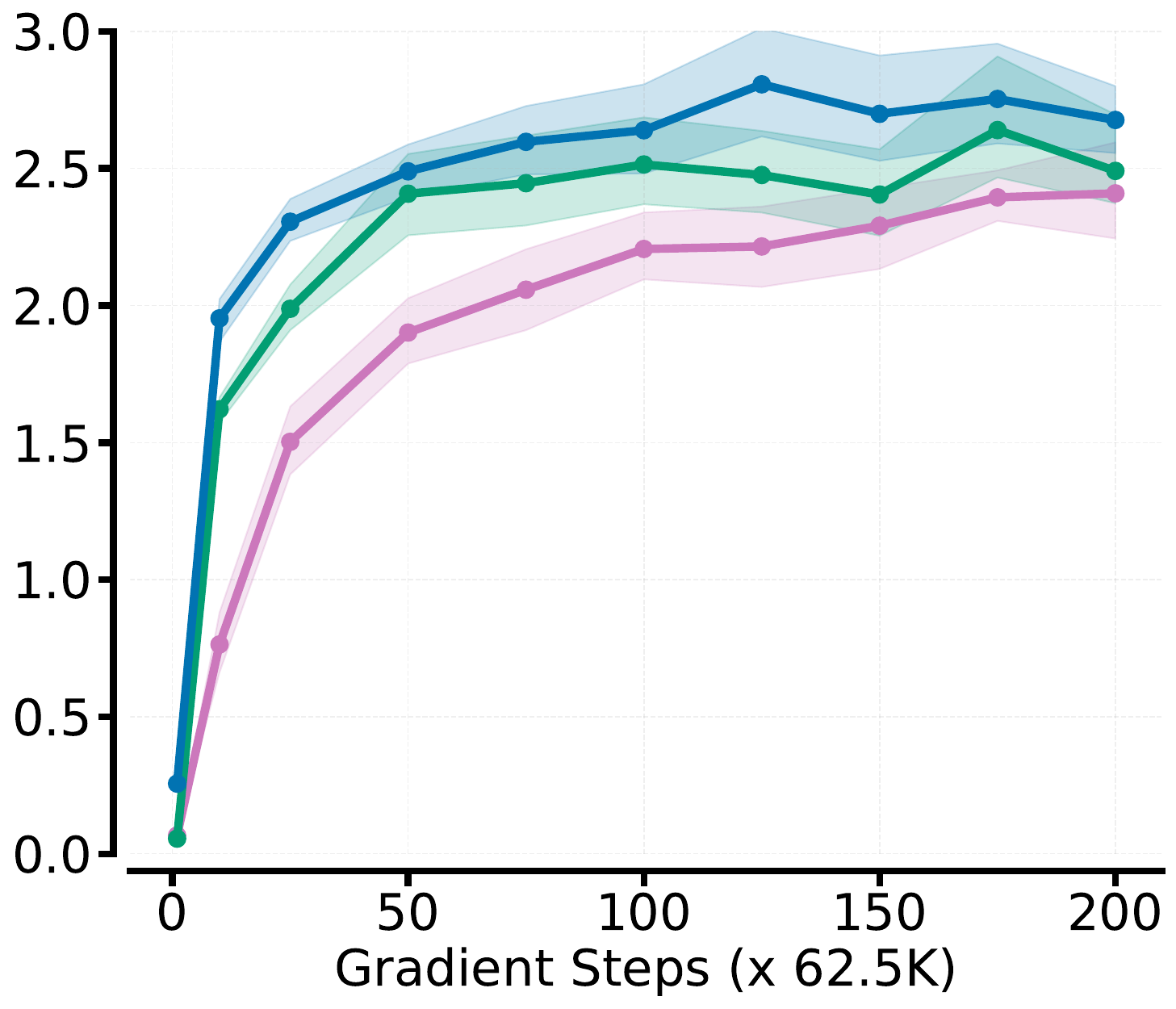}
    \caption{\textbf{Normalized performance across $17$ Atari games for CQL+C51}. x-axis represents gradient steps; no new data is collected. \textbf{Left:} $10\%$ and \textbf{Right:} $50\%$ uniform replay. We report IQM with $95\%$ stratified bootstrap CIs \citep{agarwal2021deep}}.
    \label{append:data_percentage}
\end{figure}

\newpage
\clearpage
\section{Varying Impala filter sizes}

When dealing with small models, it's common to scale them up to enhance performance. This makes the scaling strategy crucial for balancing accuracy and efficiency. For Convolutional Neural Networks (CNNs), traditional scaling methods usually emphasize model depth, width, and input resolution \citep{ding2022scaling}, as well as the \textit{filter}. The default filter size is $3$x$3$ for the Impala CNN, and we ran experiments with and without SoftMoE using $4$x$4$ and $6$x$6$ filters to investigate the filter size scaling benefits. In both cases, SoftMoE outperforms the baseline.

\begin{figure}[!h]
    \centering
    \includegraphics[width=0.4\textwidth]{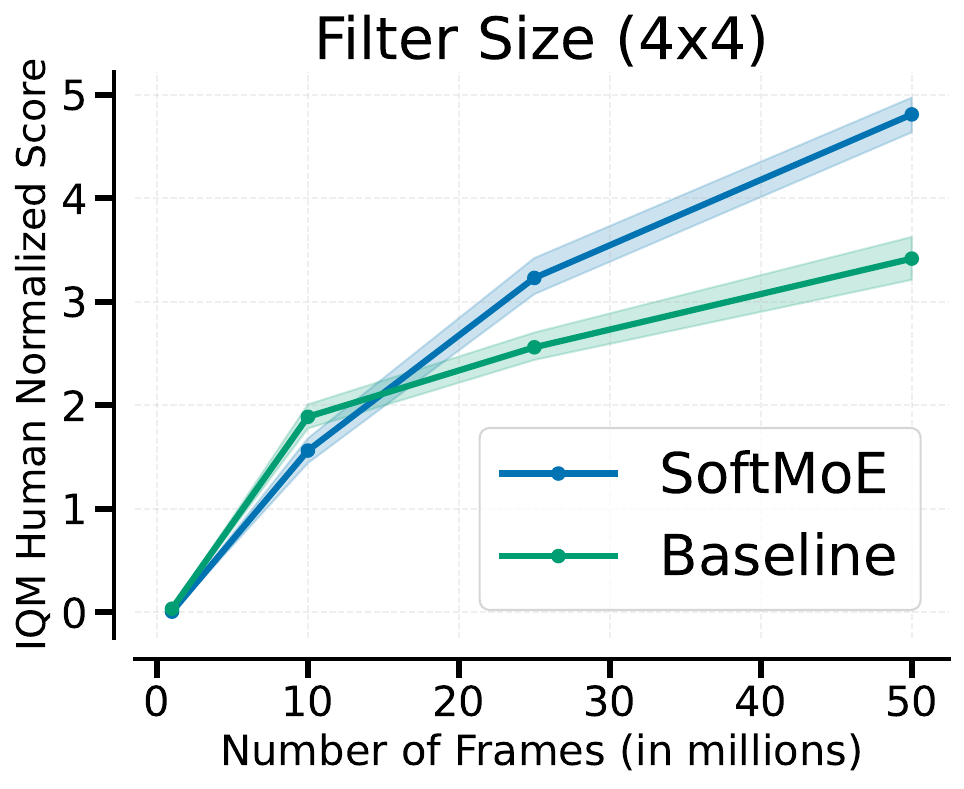}%
    \includegraphics[width=0.4\textwidth]{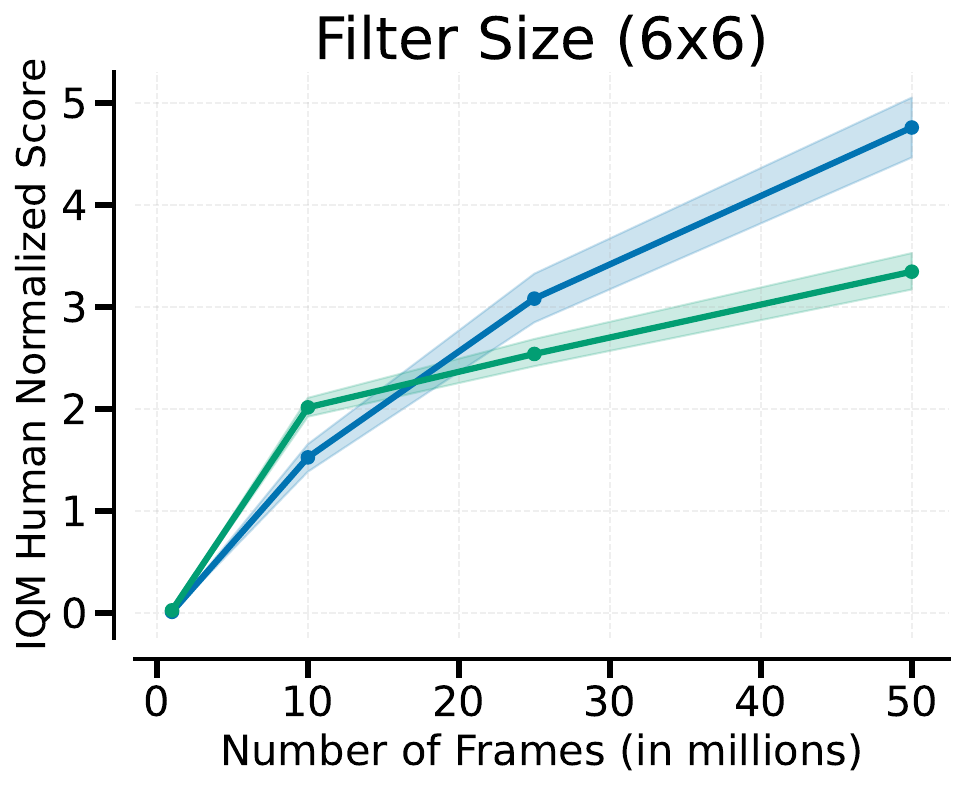}%
    \caption{\textbf{Normalized performance across $20$ Atari games with the ResNet architecture.} SoftMoE achieves the best results in both scenarios; default filter size ($3$x$3$) is increased to ($4$x$4$) and ($6$x$6$).}
\end{figure}

\section{Measuring runtime}

We plotted IQM performance against wall time, instead of the standard environment frames. SoftMoE and baseline have no noticeable difference in running time, whereas Top1-MoE is slightly faster than both.

\begin{figure}[!h]
    \centering
    \includegraphics[width=0.42\textwidth]{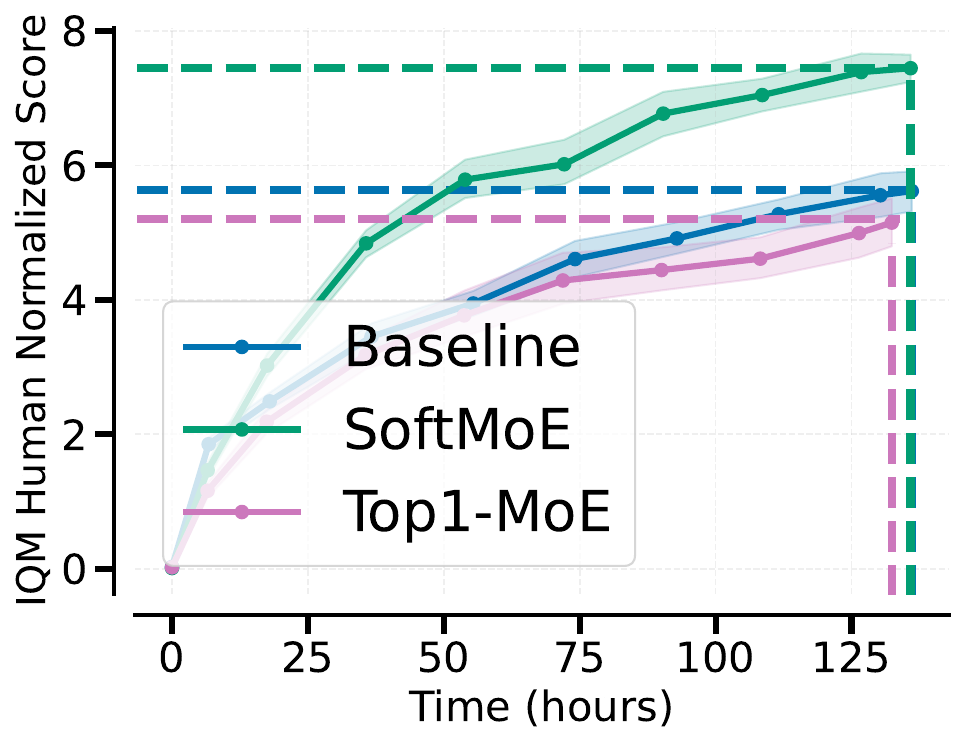}%
    \includegraphics[width=0.4\textwidth]{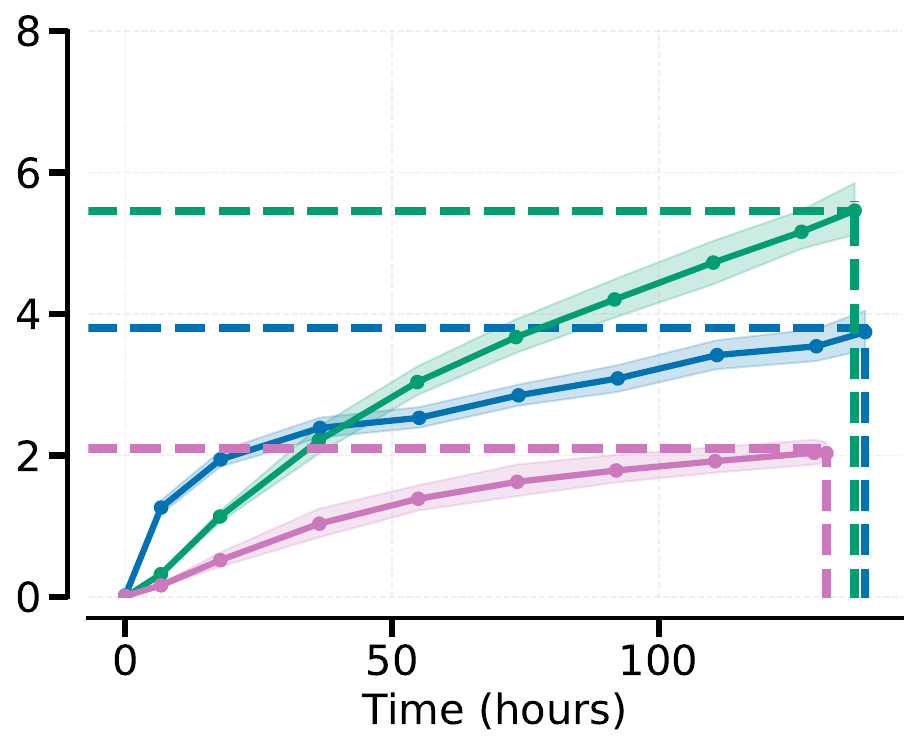}%
    \caption{\textbf{Measuring wall-time versus IQM of human-normalized scores} in Rainbow over $20$ games. \textbf{Left:} ImpalaCNN and \textbf{Right:} CNN network. Each experiment had $3$ independent runs, and the confidence intervals show $95\%$ confidence intervals.}
\end{figure}

\newpage
\clearpage
\section{Experiments with PPO}

Based on reviewer suggestions, we have run some initial experiments with PPO and SAC on MuJoCo. We have not observed significant performance gains nor degradation with SoftMoE; with Top1-MoE we see a degradation in performance, similar to what we observed in our submission. We see a few possible reasons for the lack of improvement with SoftMoE:

\begin{enumerate}
    \item For ALE experiments, all agents use Convolutional layers, whereas for the MuJoCo experiments (where we ran SAC and PPO) the networks only use dense layers. It is possible the induced sparsity provided by MoEs is most effective when combined with convolutional layers.
    \item The suite of environments in MuJoCo are perhaps less complex than the set of experiments in the ALE, so performance with agents like SAC and PPO is somewhat saturated.
\end{enumerate}


\begin{figure}[!h]
    \centering
    \includegraphics[width=0.48\textwidth]{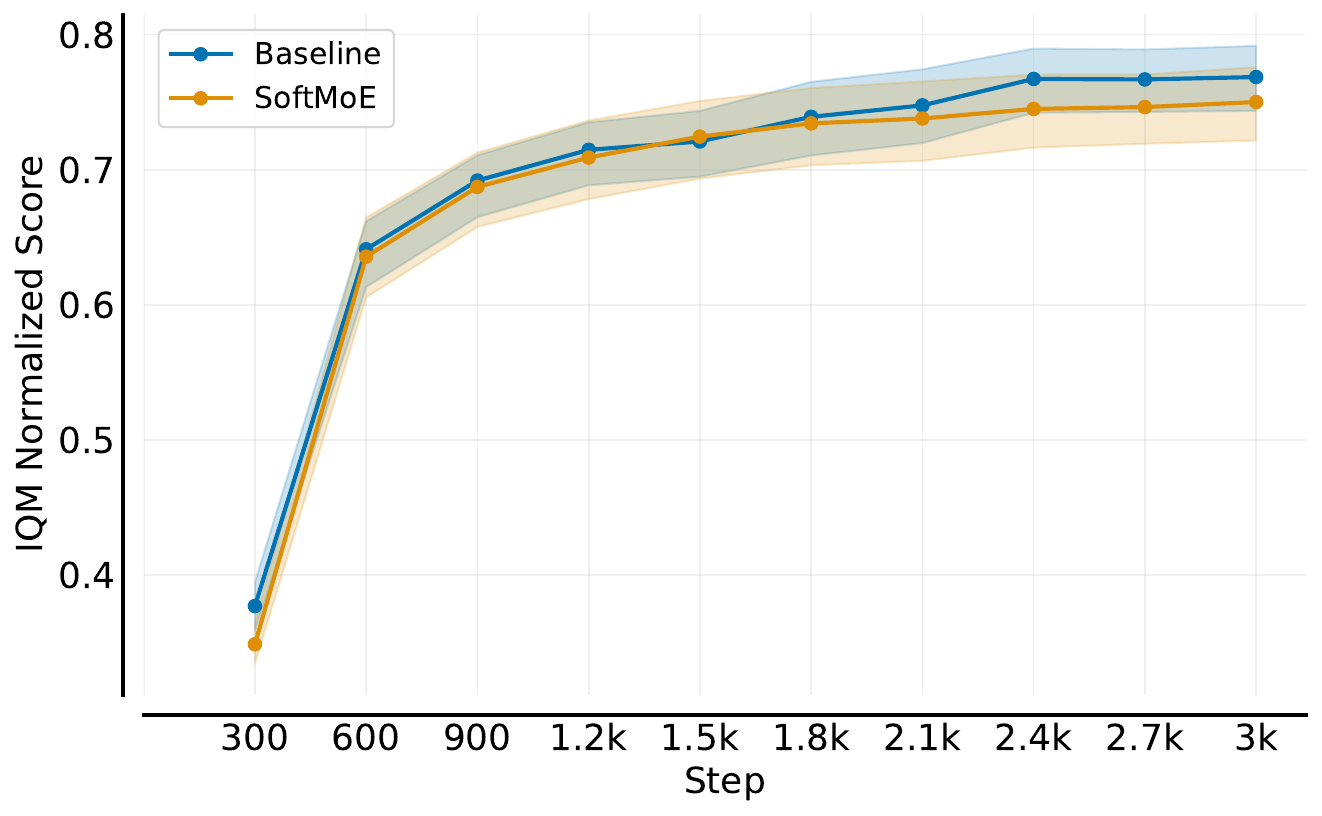} %
    \includegraphics[width=0.48\textwidth]{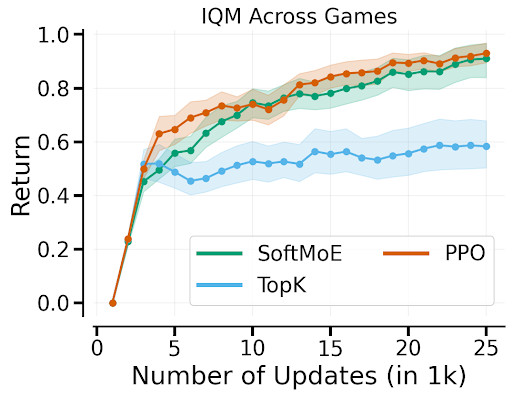}
    \caption{\textbf{Left:} Evaluating SAC with SoftMoE on $28$ MuJoCo environments and \textbf{Right:} Evaluatin PPO on $9$ MuJoCo-Brax environments. SoftMoEs seems to provide no gains nor degradation, whereas TopK seems to degrade performance (consistent with paper's findings). MuJoCo scores are normalized between $0$ and $1000$, with $5$ seeds each; error bars indicate $95\%$ stratified bootstrap confidence intervals. MuJoCo-Brax scores are normalized with respect to \citet{jesson2023relu}.}
    \label{fig:sacIQM}
    \vspace{-0.2cm}
\end{figure}


\end{document}